\documentclass[10pt,twocolumn,letterpaper]{article}

\usepackage{wacv}
\usepackage{times}
\usepackage{epsfig}
\usepackage{graphicx}
\usepackage{amsmath}
\usepackage{amssymb}
\usepackage[small]{caption}
\usepackage{multirow}
\usepackage[accsupp]{axessibility}  

\def\wacvPaperID{134} 

\wacvfinalcopy 

\ifwacvfinal
\usepackage[breaklinks=true,bookmarks=false]{hyperref}
\else
\usepackage[pagebackref=true,breaklinks=true,colorlinks,bookmarks=false]{hyperref}
\fi

\begin{document}
	
	\title{DyStyle: Dynamic Neural Network \\
		for Multi-Attribute-Conditioned Style Editings}
	
	\author{Bingchuan Li\footnotemark[1]~~~~~~~~~~~Shaofei Cai\footnotemark[1]~~~~~~~~~~~Wei Liu~~~~~~~~~~~Peng Zhang~~~~~~~~~~~\\Qian He~~~~~~~~~~~Miao Hua~~~~~~~~~~~Zili Yi\footnotemark[2]\\
		ByteDance Ltd, Beijing, China\\
		{\tt\small \{libingchuan, caishaofei, 
			liuwei.jikun, zhangpeng.ucas, heqian, huamiao, yizili\}@bytedance.com}
	}
	
	\maketitle

	
	\renewcommand{\thefootnote}{\fnsymbol{footnote}}
	\footnotetext[1]{Co-first author}
	\footnotetext[2]{Corresponding author}

	\begin{abstract}
		The semantic controllability of StyleGAN is enhanced by unremitting research. Although the existing weak supervision methods work well in manipulating the style codes along one attribute, the accuracy of manipulating multiple attributes is neglected.  Multi-attribute representations are prone to entanglement in the StyleGAN latent space, while sequential editing leads to error accumulation. To address these limitations, we design a Dynamic Style Manipulation Network (DyStyle) whose structure and parameters vary by input samples, to perform nonlinear and adaptive manipulation of latent codes for flexible and precise attribute control. In order to efficient and stable optimization of the DyStyle network, we propose a Dynamic Multi-Attribute Contrastive Learning (DmaCL) method: including dynamic multi-attribute contrastor and dynamic multi-attribute contrastive loss, which simultaneously disentangle a variety of attributes from the generative image and latent space of model. As a result, our approach demonstrates fine-grained disentangled edits along multiple numeric and binary attributes. Qualitative and quantitative comparisons with existing style manipulation methods verify the superiority of our method in terms of the multi-attribute control accuracy and identity preservation without compromising photorealism. 
	\end{abstract}
	
	\section{Introduction}
	
	Recent development in Generative Adversarial Networks (GANs) has provided a new paradigm for realistic image generation. As one of the most celebrated GAN frameworks, StyleGAN and a series of  upgraded works \cite{karras2019style,karras2020analyzing,karras2020training},  can produce diverse and high fidelity images with unmatched photorealism. Due to the introduction of scale-disentangled latent space, StyleGAN provides possibilities for flexible and controllable manipulation of attributes.
	
	\begin{figure}[t]
		\centering
		\includegraphics[width=.96\linewidth]{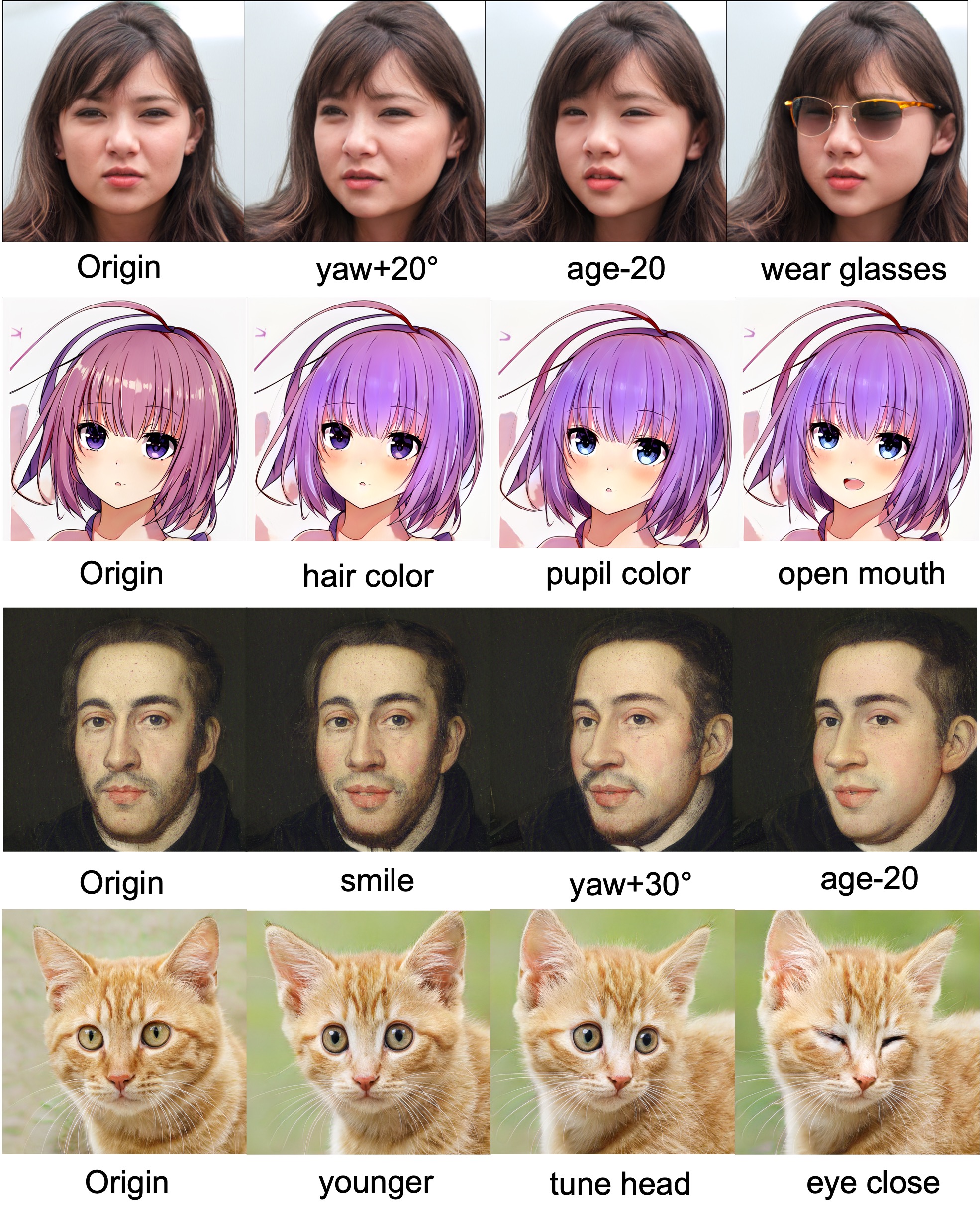}
		\captionof{figure}{The multi-attribute-conditioned image editing results with our approach, achieved by manipulating the latent codes of four different pre-trained StyleGAN2 models.}
		\label{fig:teaser}
	\end{figure}
	\vspace{4mm}
	
	Thus, one stream of research is to manipulate the latent codes of unconditional GANs, without retraining the generator. By operating the code to walk in the StyleGAN latent space, resulting in continuous changes of different degrees of a single attribute. Manipulating individual attribute of generated images has yielded satisfactory results \cite{shen2020interfacegan,abdal2021styleflow,wang2021hijack}. However, performing edits along one attribute may lead to unexpected changes in other semantics due to entanglement between different semantics in the latent space. Prior methods always produce semantic confusion when jointly manipulating multiple attributes. We analyzed several reasons: 1. The static structure of single-attribute sequential editing is prone to error accumulation. E.g. InterfaceGAN \cite{shen2020interfacegan}, the identity similarity drops sharply as the number of edited attributes increases;  2. The static structure of multi-attribute parallel \cite{liu2021isf,abdal2021styleflow} input fails to adapt to random combinations of attributes during inference. Since the structure and parameters are fixed after training, all attributes must be calculated forward regardless of whether they are selected or not; 3. Training parallel static structures requires simultaneously optimizing all attributes in one forward pass, which makes learning disentangled latent space more difficult.
	
	We argue that the style editing network should be able to adapt to wide varieties of attribute configurations, and when training the style editing network for multiple attribute manipulation, any biases of the distribution of attribute configurations would easily result in systematic control errors. Take realistic portrait editing as an example, for one case we only change the hair color of the portrait. For another case, we change both the hair color and age of the portrait. We wish the two attribute configurations are sampled with equal probability and are both well handled by the style editing network. We made two efforts to address this issue, we evenly sample the attribute configurations during training instead of using static set of training samples. To make this possible, we employ pre-trained knowledge networks to provide “on-the-fly” supervisions rather than using static labels. In addition, we employ a dynamic style editing network consisting of multiple experts, each of which is responsible for the manipulation of one attribute. We dynamically activate a subset of the experts based on whether the corresponding attributes are edited or not. That means, the structure and parameters of the style editing network vary by different input samples. Experiments show that the Dynamic Style Manipulation Network (DyStyle) that processes each sample with data-dependent architectures and parameters can well adapt to various types of attribute configurations.
	
	Although dynamic networks are usually more flexible than static networks, they also require appropriate optimization methods to play its performance. Therefore, we propose dynamic multi-attribute contrastive learning (DmaCL) methods cooperate with better optimization of DyStyle. In general, we perform contrastive constraints from the image space and latent space of model. For generated images, we employ a variety of pre-trained classifications or regressions to capture related attribute features. Dynamic multi-attribute contrastor are proposed to ensure the other attributes of the image are unchanged when optimizing activated attributes. Meanwhile, we propose dynamic multi-attribute contrastive loss to constrain the embeddings extracted by the expert network. It can perceive and discriminate the correlation and difference between the activated attributes to ensure disentanglement during training. To make the training of the DyStyle network easier, we adopt a novel easy-to-hard training procedure in which the DyStyle network is trained for editing a single attribute at a time, and then trained for jointly manipulating multiple randomly-sampled attributes. Generally speaking, our contributions include:
	
	\begin{itemize}
		\item{A Dynamic Style Manipulation Network (DyStyle) is carefully designed to perform multi-attribute-conditioned editing of the StyleGAN latent codes,  implementing adaptability to wide varieties of attribute configurations.	}
		\item{We propose the Dynamic Multi-Attribute Contrasive Learning (DmaCL) method, which is applied to fully optimize the dynamic neural network to realize the disentanglement of image space and latent space. }
		\item{Comprehensive evaluations on various datasets (realistic faces, comics, artistic portraits, animal faces) demonstrate improved attribute control accuracy and better identity preservation of our approach over existing static architectures. The improvements are more significant when jointly manipulating the styles along multiple attributes.}
	\end{itemize}
	
	
	\section{Ralated Work}
	
	\subsection{Unconditional GANs}
	
	Generative Adversarial Network (GAN) is first introduced by Goodfellow et al. \cite{goodfellow2014generative}, and has been one of the most active fields in deep neural networks. One research direction is to improve the GAN architectures, loss functions, and training dynamics for improved quality, diversity and stability of training. In terms of the GAN architectures, DCGAN \cite{radford2015unsupervised}, ProgressiveGAN \cite{karras2017progressive}, BigGAN \cite{brock2018large} and StyleGAN series \cite{karras2019style,karras2020analyzing,Karras2020ada,Karras2021} architectures are the top known architectures in history of development. We build our work on StyleGAN2 \cite{Karras2020ada}, as it achieves the best photorealism.
	
	\begin{figure*}[t]
		\centering
		\includegraphics[width=0.95\linewidth]{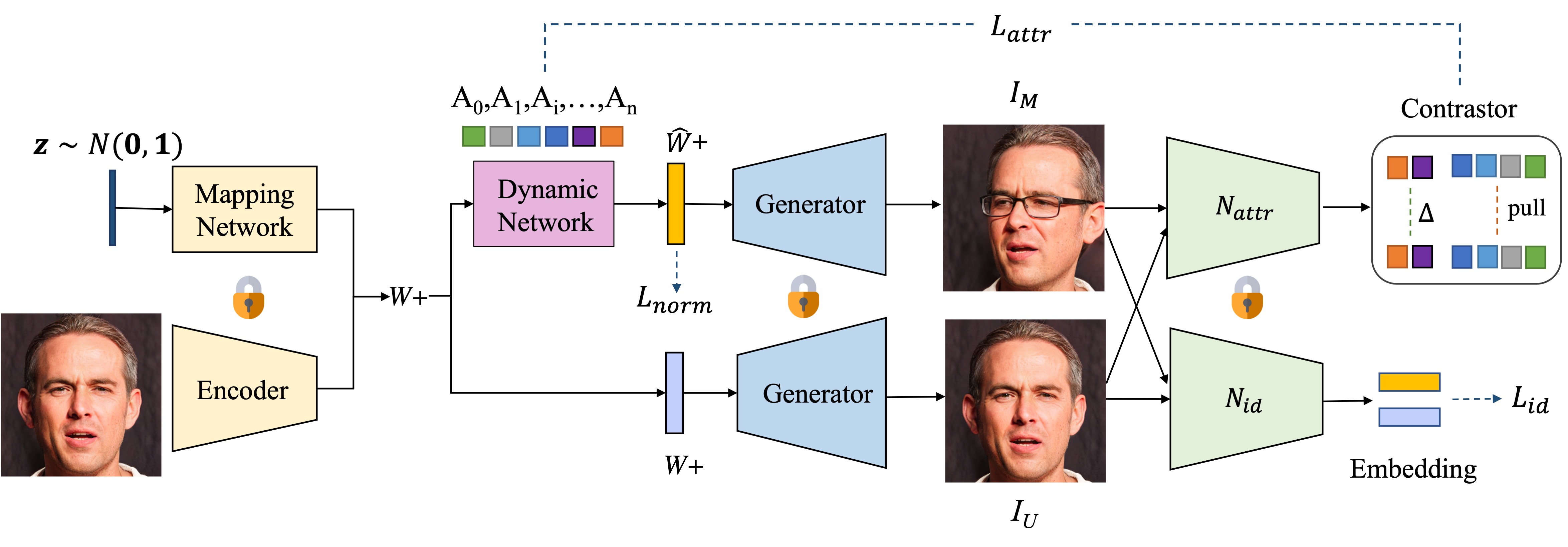}
		\caption{The framework and training losses of our multi-attribute-conditioned style editing approach. The Dynamic Network is trained for attribute-conditioned image editing while the StyleGAN2 Generator, the Style Mapping Network and the Encoder are held fixed.}
		\label{fig:framework}
	\end{figure*}
	\subsection{Attribute Conditional GANs}
	
	Conditional GANs \cite{mirza2014conditional} have given rise to many image manipulation applications. Unlike unconditional GANs that take random noises as input, conditional GANs take meaningful priors (e.g., attribute \cite{liu2021isf,abdal2021styleflow}, text \cite{patashnik2021styleclip,xia2021tedigan,wang2021hijack}, 3D model parameters \cite{tewari2020stylerig}, sketches \cite{zeng2022sketchedit,wang2021sketch}, label maps \cite{park2019SPADE,shi2022semanticstylegan}) as input and synthesizes relevant images, thus offering users a certain level of control. 
	
	Some of unsupervised methods \cite{harkonen2020ganspace,shen2020closed} uses Principal Component Analysis (PCA) to learn the most important directions. Although these methods can find some principal components with clear semantics, they need to be artificially defined. Unlike unsupervised methods, supervised attribute conditional GANs usually associate attribute labels to GAN latent space to manipulate image generation. InterFaceGAN \cite{shen2020interfacegan} uses labeled data to learn a SVM to discover the separation plane and the directions of certain attributes. StyleRig \cite{tewari2020stylerig} employs a 3D morphable face models (3DMMs) \cite{blanz1999morphable}, and a rigging network to map 3DMM’s semantic parameters to StyleGAN’s input. Although StyleRig generates very nice results for the manipulation of head pose and illumination, the detailed control of other facial attributes did not work. StyleFlow \cite{abdal2021styleflow} seeks continuous and nonlinear normalizing flows in the latent space conditioned by attribute features. However, the training of StyleFlow does not explicitly enforce the control accuracy of attributes, and the assumption of normal distribution with respect to any attribute configuration does not always hold due to data biases. StyleCLIP \cite{patashnik2021styleclip} introduces text-aware CLIP \cite{2021Learning} to generative models, thereby enriching manipulable semantics. ISFGAN \cite{liu2021isf} proposes an implicit style function to straightforwardly achieve multi-modal and multi-domain image-to-image translation from pre-trained unconditional generators.
	

	\section{Method}
	\subsection{Framework}
	\begin{figure*}[t]
		\includegraphics[width=0.95\linewidth]{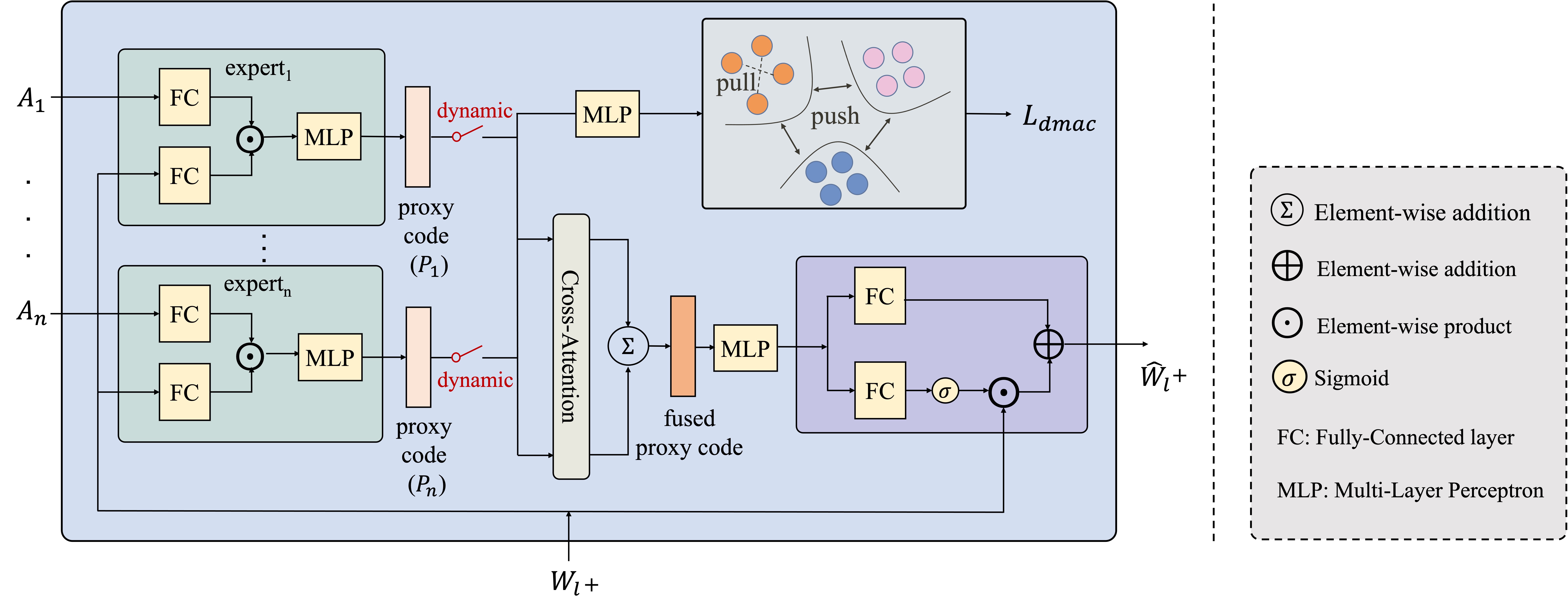}
		\centering
		\caption{The architecture of our Dynamic Style Manipulation Network (DyStyle). $\textbf{W}_l+$ ($l\in\{0,1,...,17\}$) is one of the 18 vectors of $\textbf{W}+$. Multiple experts are employed, in which each expert is responsible for the processing of one attribute before they are joined. Whether an expert is activated is based on whether the corresponding attribute is intended for editing or not.}
		\label{fig:dnn}
	\end{figure*}
	As shown in Figure \ref{fig:framework}, our method manipulates the extended latent code $\textbf{W}+$ using a Dynamic Style Manipulation Network (DyStyle). The extended latent code $\textbf{W}+$ consists of 18 different 512-dimensional vectors, one for each input layer of StyleGAN2 generator~\cite{karras2020analyzing}. The $\textbf{W}+$ can be either mapped from a random Gaussian noise vector $\textbf{z} \sim N(\textbf{0}, \textbf{1})$  with the Style Mapping Network of StyleGAN2 \cite{karras2020analyzing} or embedded from a real photograph with the image-to-style encoder of pSp \cite{richardson2020encoding} or E4E \cite{tov2021designing}. The DyStyle network takes an attribute specification and $\textbf{W}+$ as inputs and predicts a manipulated latent code $\textbf{\^{W}}+$. The attribute specification $\textbf{Attr}$ is made up of a set of attribute values specified by the user, defining the appearance of the desired image. The attribute set, in our experimental setting, includes numeric attributes (e.g., yaw, pitch of the head pose, age) and binary attributes (e.g., glasses, smile,  black hair, mustache, close eye, open mouth). The number of attributes can be expanded without modifying the framework.
	
	The manipulated latent code $\textbf{\^{W}}+$ is fed to the StyleGAN2 generator to generate the corresponding manipulated image $\textbf{I}_M$. In the meantime, the original latent code $\textbf{W}+$ is mapped to the untouched image $\textbf{I}_U$. $\textbf{I}_M$ is expected to maintain the identity of $\textbf{I}_U$, while matching the target attribute specification. Dynamic multi-attribute contrastor here is utilized to learn the disentanglement between activated and inactive attributes. Such constraint is enforced with the pre-trained attribute predictors $N_{attr}$ and a pre-trained identity recognition model $N_{id}$.
	
	\subsection{Dynamic network architecture}
	\label{sect:architecture}
	
	The architecture of DyStyle is shown in Figure \ref{fig:dnn}.  As shown, the DyStyle network manipulates each $\textbf{W}_l+$ code separately, by taking the attribute configuration and $\textbf{W}_l+$ as input and predicting the proxy code $P$ which is further used to linearly modulate the $\textbf{W}_l+$ code itself. By conditioning the proxy code jointly upon the attribute configuration and $\textbf{W}_l+$, rather than the attribute configuration solely,  the network is able to predict the proxy code adaptively for each input case of  $\textbf{W}+$, rather than generate a uniform modulation parameters for all cases of $\textbf{W}+$. The DyStyle network employs multiple experts to process the attributes separately before they fused with cross-attention and element-wise addition. The proxy codes extracted by activated expert are encoded into a unified latent space, so that the dynamic multi-attribute contrastive loss is performed to constrain the same attribute aggregation and different attributes to separate from each other. Cross-attention allows different experts to communicate with each other and enables an expert adapt to the unexpected influences caused by the edits of other attributes. This argument is well supported by the ablation studies: see the Sec. 6.2 for more details. Addtionally, the cross-attention module is well suited for a variable number of attributes, which is an important feature of our dynamic architecture.
	
	We join the features after attribute-specific processing by cross-attention and element-wise addition. Such a design favors disentangled attribute editing and improved control precision. The cross attention is computed as Eq. \ref{eq:crossattention}.
	\begin{equation}
		\footnotesize
		\label{eq:crossattention}
		P_i =   \sum_j  V_j \odot \frac{\exp(Q_i \cdot K_j)}{\sum_j \exp(Q_i \cdot K_j) } 
	\end{equation}
	where $Q_i, K_i, V_i$ $i \in \{1,2,...,n\}$are the query, key and value vectors computed from the proxy code $P_i$ with an FC layer respectively. As an extreme case, when $n=1$, the output of the cross-attention $ P_i =   V_i$.
	
	
	\subsection{Dynamic multi-attribute contrastive learning}
	
	The DyStyle network is trained with an objective consisting of 4 types of losses, which is defined as
	\begin{equation}
		\footnotesize
		\label{eq:loss}
		L =  \alpha_{attr} L_{attr} + \alpha_{dmac} L_{dmac}  + \alpha_{id} L_{id} +  \alpha_{norm} L_{norm} 
	\end{equation}
	where $L_{attr}$ is the attribute loss for various attributes (e.g., pose, age, black hair, glasses, smile), which are controlled by a dynamic multi-attribute contrastor and used to enforce the consistency of target attributes and measured attributes of the manipulated image $\textbf{I}_M$. $ L_{dmac}$ is designated to disentangle the activated attributes in the latent space of the model. $L_{id}$ is the identity loss intended to preserve the identity of the original image, while $L_{norm}$ is the normalization loss discouraging degradation of image quality. $\alpha_{attr}$, $\alpha_{dmac}$, $\alpha_{id}$ and  $\alpha_{norm}$ are the corresponding coefficients for each loss term. 
	
	\subsubsection{Dynamic multi-attribute contrastor}
	\label{sect:loss}
	Similar to some weakly supervised methods \cite{alaluf2021matter,patashnik2021styleclip}, we employ pre-trained models to extract attribute information. Since the training mode of the framework relies on dynamic attribute input, the goal of dynamic multi-attribute contrastor is to ensure the generated image after manipulated contains activated attributes, while the inactive attributes remain consistent with randomly generated images. The form of  $L_{attr}$ that relies on a set of contrastors differs for numeric attributes and binary ones. We design specific losses for different types of pre-trained estimators.
	
	Specifically, the contrastive loss for numeric attribute $A^k$ is defined as
	\begin{equation}
		\footnotesize
		L_{attr}^{A^k}=
		\begin{cases}
			\left | A^k_M - A^k_U\right |, \text{if $\Delta^{gt}_{A^k}$ is none}\\
			\max(\left | A^k_M - A^k_U - \Delta^{gt}_{A^k}\right | - T_{A^k}, 0 ), \text{otherwise}
		\end{cases}
	\end{equation}
	where $A^k_M $, $A^k_U$ are attribute value of Face $\textbf{I}_M$, $\textbf{I}_U$ measured by a pre-trained attribute estimation network $N_{A^k}$. $\Delta^{gt}_{A^k}$ is the ground-truth pose variation of $\textbf{I}_M$ to $\textbf{I}_U$, specified at the input of DyStyle. If $\Delta^{gt}_{A^k}$ is none, it means ${A^k}$ is inactive. $T_{A^k}$ are constant thresholds, which is set to 3 for yaw and pitch, and 5 for age.
	
	For binary attributes, the input attribute values are either 0 or 1, and they represent the status of the target attributes (1 means ``with’’ and 0 means ``without’’). We employ a pre-trained multi-task or multi-class network ($N_{A^k}$) to predict all the binary attributes of an image, thus $\textbf{A}^k_M=N_{A^k}(\textbf{I}_M)$, $\textbf{A}^k_U=N_{A^k}(\textbf{I}_U)$. 
	
	The attribute loss of the binary attributes is written as
	\begin{equation}
		\footnotesize
		L^{A^k}_{attr} =
		\begin{cases}
			1 - \frac{\textbf{emb}^{A^k}_M}{\|\textbf{emb}^{A^k}_M\|} \cdot \frac{\textbf{emb}^{A^k}_U}{\|\textbf{emb}^{A^k}_U\|}, \text{if $\textbf{A}^k_{gt} == \textbf{A}^k_U$ or $\textbf{A}^k_{gt}$ is None} \\
			- \sum_{A^k} [A^k_{gt} \log A^k_M  + (1-A^k_{gt}) \log (1-A^k_M)], \text{otherwise}
		\end{cases}
	\end{equation}
	where $\textbf{emb}^{A^k}_M$ (or $\textbf{emb}^{A^k}_U$) is the activation of the second last layer of the pre-trained multi-attribute predictor $N_{A^k}$ given $\textbf{I}_M$ (or $\textbf{I}_U$) as input. $\textbf{A}^k_{gt}$ is the ground-truth (or target) attribute specification. The similarity score of $\textbf{emb}^{A^k}_M$ and $\textbf{emb}^{A^k}_U$ is enforced to 1 when no edits are intended. Further, the cross-entropy loss is calculated separately for each binary attribute, and finally summed up. 
	
	The original face $\textbf{I}_U$ and the manipulated face $\textbf{I}_M$ are expected to have the same identity. Therefore, the identity loss are computed as
	\begin{equation}
		\label{eq:lpips}
		L_{id} = 1 - cos\_similarity(\textbf{emb}^{id}_M, \textbf{emb}^{id}_U)
	\end{equation}
	where $\textbf{emb}^{id}_M$ (or $\textbf{emb}^{id}_U$) is the feature embedding of the face in $\textbf{I}_M$ (or $\textbf{I}_U$), extracted with a pre-trained face recognition model. When the editing targets are not realistic faces, the conventional face recognizer does not serve as an effective identity representation. In this cases (e.g., comics or animal faces), we employ LPIPS loss \cite{zhang2018unreasonable} as the identity loss. 
	
	Practically, we choose $\alpha_{yaw}=0.05$, $\alpha_{pitch}=0.05$,  $\alpha_{age}=0.02$,  $\alpha_{A^k}=1.0$ (if $A^k$ is binary attribute), $\alpha_{id}=1.0$ and $\alpha_{norm}=0.001$  in our experiments. Not that the hyperparameters are adjusted to scale the different types of losses to the same magnitude.
	
	\subsubsection{Dynamic multi-attribute contrastive loss}
	\label{sect:dmacl}
	\begin{figure*}[t]
		\centering
		\includegraphics[width=\linewidth]{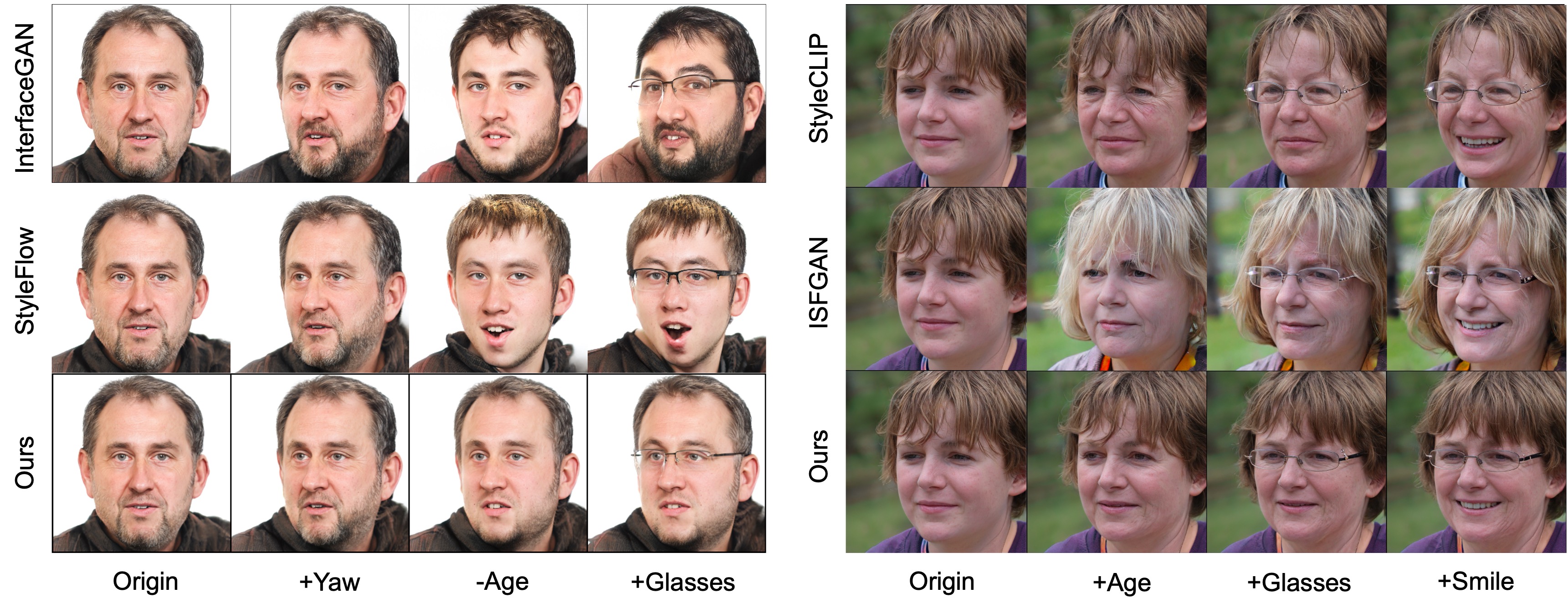}
		\caption{Comparisons of our method and competing methods in terms of multiple-attribute manipulation. As shown, the compared methods saw significant identity variation and semantic entanglement.}
		\label{fig:comp}
	\end{figure*}
	
	The proposed dynamic model will randomly select attributes for activation in a training iteration. The dynamic multi-attribute contrastor can capture the difference between the active and inactive attributes during training. However, a variety of attributes come from different pre-trained estimators. It is difficult for the contrastor to disentangle the attributes that are activated at the same time. Therefore, we apply contrastive loss $L_{dmac}$ on the simultaneously activated attributes in the latent space of the model. The principle of the $L_{dmac}$ is to ensure that the editing directions of a single attribute is fully gathered, while increasing difference between the editing directions of multiple attributes as much as possible.
	
	As shown in Figure \ref{fig:dnn}, attributes are independently encoded by experts as proxy codes, which contains the direction of attribute changes. In order to perform contrastive learning, activated proxy codes are mapped into the unified latent space. Here, the encoder simply applies a Multilayer Perceptron (MLP). We calculate the attribute auto-correlation and cross-correlation respectively. 
	For auto-correlation, we calculate it on the batch dimension which the similarity measured between attributes by dot product. The formula is
	\begin{equation}
		\label{eq:ac}
		I^{k}_{ac} = \sum_{i,j\in N_{b}, i\neq j}S_{i}^{k}\cdot S_{j}^{k}
	\end{equation}
	where $N_{b}$ is the training batch size, and $S_{i}^{k}$ represents the i-th dimention of the k-th attribute code in the unified latent space.
	For the cross-correlation coefficient, the attribute codes other than itself are regarded as negative samples, so the formula is
	\begin{equation}
		\label{eq:cc}
		I^{k}_{cc} = \sum_{q\in N_{k}, q\neq k}S^{k}\cdot S^{q}
	\end{equation}
	where $N_{k}$ represents that k attributes are activated by the current dynamic network. The form of our function $L_{dmac}$ follows InfoNCE \cite{oord2018representation}, which is calculated as
	\begin{equation}
		\label{eq:dmac}
		L_{dmac}=\sum_{k\in N_{k}}-\log \frac{\exp\left(I^{k}_{ac}\right)}{\exp\left(I^{k}_{ac}\right)+\exp\left(I^{k}_{cc}\right)}
	\end{equation}
	Note that each activation attribute is traversed in a single training iteration to calculate the loss sum. The form of $L_{dmac}$ can be easily realized using softmax cross-entropy loss.
	
	For $\textbf{\^{W}}_l+$ output by the dynamic network, the normalization loss is defined as
	\begin{equation}
		\label{eq:norm}
		\begin{split}
			L_{norm} =  \sum_l \| (\textbf{\^{W}}_l+) - \textbf{W}_{avg} \|
		\end{split}
	\end{equation}
	where $\textbf{W}_{avg}$ is the statistic center of the $\textbf{W}$ space of the pre-trained StyleGAN2 generator \cite{karras2020analyzing}, and $\textbf{\^{W}}_l+$ is the manipulated style vector corresponding to the \textit{l}-th layer. As discussed in \cite{richardson2020encoding}, being closer to $\textbf{W}_{avg}$ means higher expected quality of the generated image.
	
	\subsubsection{Two-stage training strategy}
	\label{sect:twostage}
	
	The DyStyle is trained with randomly sampled $\textbf{W}+$ codes with the Style Mapping Network and evenly-sampled attribute configurations. The training is conducted by following a two-phase procedure. In this first stage, the network is trained for single-attribute manipulation, by randomly choosing an attribute for editing and evenly sample the target attribute value. That means, only an expert (or branch) for the edited attribute is activated at a time in this phase. Note that the loss is not changed. This phase allows each expert to focus on one attribute at a time and get used to easy editing cases. In the second stage, the DyStyle network is trained to adapt to situations when multiple attributes are manipulated jointly. In this phase, a combination of attribute set are randomly sampled and set as the input of the DyStyle network, so that the experts learn to communicate with each other and adapt to more complex attribute configurations.

	\section{Experimental Results}
	
	\subsection{Experiment setups}
	\begin{table*}[t]
		\centering
		\caption{Quantitative comparisons of different attribute-conditioned style editing approached in terms of the identity preservation and attribute control accuracy. Note that we compute Mean Absolute Error for numeric attributes (e.g., yaw, age) and Classification Acc. for binary attributes (e.g., glasses, smile). The identity similarity score is between the original face and the manipulated.}
		\label{table:comp}
		\resizebox{\textwidth}{!}{
			\begin{tabular}{ l||c|c|c|c|c|c|c|c } 
				&   \multicolumn{4}{c|}{single-attribute editing} & \multicolumn{4}{c}{multi-attribute editing}      \\ 
				{\color{white}wwwwww}attribute type& yaw & age &  glasses &smile&  \multicolumn{2}{c|}{glasses+smile} & \multicolumn{2}{c}{yaw+glasses}      \\ 
				\hline
				\hline
				method& \multicolumn{8}{c}{Identity Similarity Score($\uparrow$)}\\
				\hline
				InterFaceGAN (\cite{shen2020interfacegan})  & 0.78 ± 0.05 & 0.82 ± 0.06 & 0.84 ± 0.12&0.95±0.05 & \multicolumn{2}{c|}{0.74 ± 0.05} & \multicolumn{2}{c}{0.68± 0.15} \\  
				StyleFlow (\cite{abdal2021styleflow})& 0.82 ± 0.07 & 0.86 ± 0.08&0.85 ± 0.1&0.96 ± 0.05&\multicolumn{2}{c|}{0.83 ± 0.12} &\multicolumn{2}{c}{0.78 ± 0.12} \\
				ISFGAN \cite{liu2021isf} & - & 0.85 ± 0.13&0.88 ± 0.03&0.96± 0.03&\multicolumn{2}{c|}{0.85 ± 0.08} &\multicolumn{2}{c}{-} \\
				Ours (w/o $L_{id}$)  & 0.85 ± 0.04 & 0.86 ± 0.03 & 0.87  ± 0.08 & 0.97 ± 0.02 & \multicolumn{2}{c|}{0.85 ± 0.12} &\multicolumn{2}{c}{0.82 ± 0.24} \\ 
				\textbf{Ours} (w/ $L_{id}$)   & \textbf{0.95} ± 0.05 & \textbf{0.89} ± 0.08 & \textbf{0.90}  ± 0.09 & \textbf{0.98} ± 0.1 & \multicolumn{2}{c|}{\textbf{0.87} ± 0.1} &\multicolumn{2}{c}{\textbf{0.85} ± 0.09} \\  
				\hline
				& \multicolumn{8}{c}{Attribute Control Accuracy}\\
				method& yaw($\downarrow$) & age ($\downarrow$) & smile ($\uparrow$) & glasses ($\uparrow$) &glasses($\uparrow$)&smile($\uparrow$)&yaw($\downarrow$)&glasses($\uparrow$)\\
				\hline
				InterFaceGAN (\cite{shen2020interfacegan}) & 12.95& 13.50& 0.894 & 0.93 & 0.832& 0.877 & 15.33& 0.826 \\  
				StyleFlow (\cite{abdal2021styleflow})& 6.41&\textbf{12.78}&0.975 & 0.981 & 0.944& 0.921& 8.58  & 0.925\\  
				ISFGAN \cite{liu2021isf}   & - &13.97&0.963 & 0.985 & 0.935& 0.901&  -  & - \\  
				\textbf{Ours}   & \textbf{6.33}& 13.77& \textbf{0.976} & \textbf{0.988}& \textbf{0.963}& \textbf{0.955}& \textbf{7.26}  & \textbf{0.961}\\  
		\end{tabular}}
	\end{table*}
	
	\begin{figure*}[h]
		\centering
		\includegraphics[width=\linewidth]{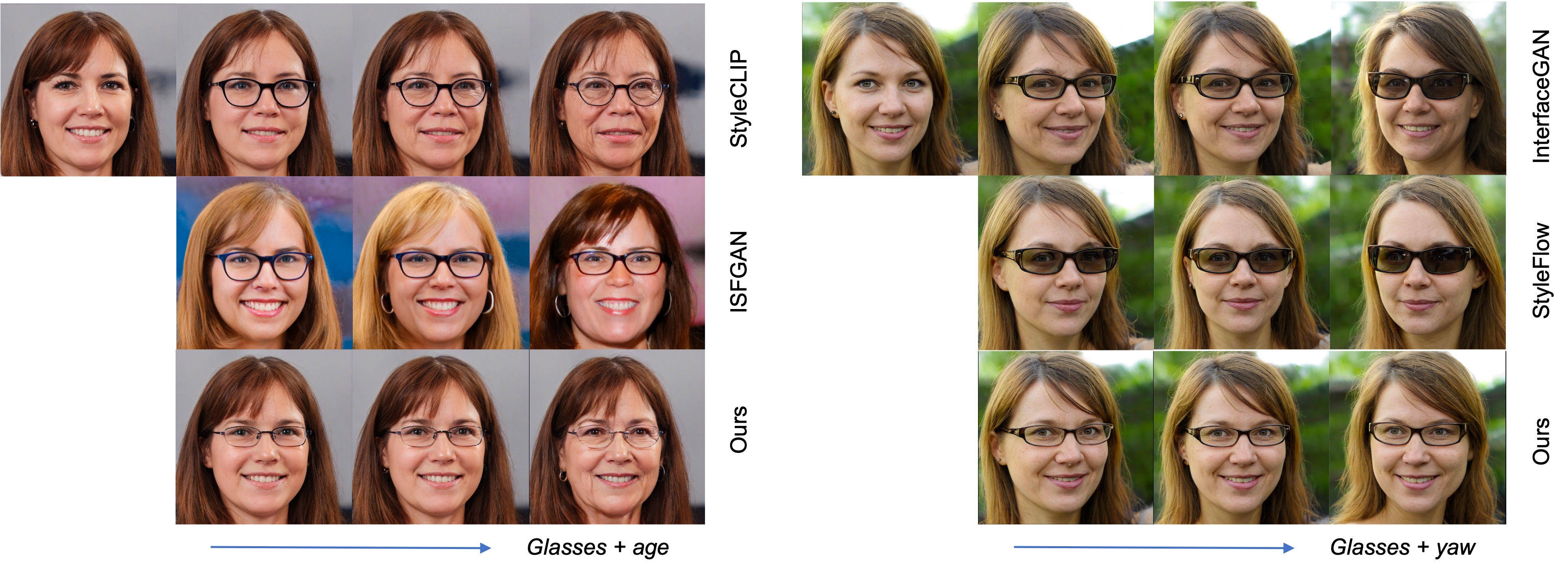}
		\caption{Fixing a specific property while continuously manipulating other properties. Glasses are chose to fix due to their instability. As shown, our method stably manipulates pose and age attributes while minimizing changes in glasses shape and color. Furthermore, our method clearly has the best consistency with the original image for unedited attributes, such as smiling expressions.}
		\label{fig:comp2}
	\end{figure*}
	
	We experimented our style manipulation network on four pre-trained StyleGAN2 models, the one trained on FFHQ dataset \cite{karras2019style} for realistic face generation, the one trained on MetFace \cite{karras2020training} for artistic face synthesis, that trained on AFHQ \cite{choi2020stargan} for animal face synthesis and the one trained on comics dataset \cite{branwen2015danbooru2019} for comic face generation. The expanded experimental setups are reflected in the supplementary materials. 
	
	Various attribute manipulation results of our approach are presented in Figure \ref{fig:teaser}. Some more attribute-controlled image generation results for each datasets are presented in Figure 15 (realistic faces), Figure 16 (artistic faces), Figure 18 (Comics) and Figure 19 (animal faces) in the supplementary materials. Some more high-resolution (1024$\times$1024) realistic face editing results by our approach are presented in Figure 17 in the supplementary materials. With the image-to-style encoder provided in \cite{richardson2020encoding}, we also conducted attribute-conditioned editing of real photos and present the results in Figure 20 in the supplementary materials.

	\subsection{Comparisons}
	
	To verify the effectiveness of the proposed method, we compared our approach with the state-of-the-art style manipulation methods including InterFaceGAN \cite{shen2020interfacegan}, StyleFlow~\cite{abdal2021styleflow}, StyleCLIP \cite{patashnik2021styleclip} and ISFGAN \cite{liu2021isf}. To assure fair comparisons, these methods are designated to manipulate the latent space of the same pre-trained StyleGAN2 model which was trained on FFHQ dataset \cite{karras2019style}.  According to the common attributes of the official open source of different methods, the comparisons only account for the editing of four attributes (yaw, age, glasses and smile).  StyleCLIP \cite{patashnik2021styleclip} and ISFGAN \cite{liu2021isf} are compared separately due to their lack of ability to edit poses. As the separation plane of InterFaceGAN only specifies the directions of attribute control, we toughly estimated  the physical meaning of scales based on a few labeled examples. Note that StyleCLIP \cite{patashnik2021styleclip} does not participate in quantitative evaluation, as CLIP-encoded \cite{2021Learning} textual information cannot be quantified consistently across scales.
	
	\setlength{\tabcolsep}{6.2pt}
	\begin{table}[b]
		\begin{center}
			\caption{The FID comparisons between manipulated faces $I_M$.}
			\label{table:comp_fid}
			\begin{tabular}{l|c|c|c|c}
				\hline\noalign{\smallskip}
				\multirow{2}*{Methods} & \multicolumn{4}{c}{FID($\downarrow$)}\\
				\cline{2-5} 
				\noalign{\smallskip}
				~ & age & glasses & smile & avg\\
				\hline
				\noalign{\smallskip}
				InterFaceGAN \cite{shen2020interfacegan} & 64.32 & 62.66 & 57.91 & 61.63\\
				StyleFlow \cite{abdal2021styleflow} & 53.5 & 51.85 & 51.34 & 52.23\\
				StyleCLIP \cite{patashnik2021styleclip} & 51.23 & 47.45 & 47.54 & 48.74\\
				ISFGAN \cite{liu2021isf} & 52.87 & 51.3 & 47.78 & 50.65\\
				\hline
				\noalign{\smallskip}
				Ours & 47.73 & 42.69 & 41.55  & \bf 43.98\\
				\hline
			\end{tabular}
		\end{center}
	\end{table}
	\setlength{\tabcolsep}{1.4pt}
	
	With the same set of 5000 attribute-configuration-and-$\textbf{W}+$ pairs as test set, we conduct quantitative evaluations on these methods. Specifically, we evaluate the accuracy (or precision) of attribute control, preservation of identity and image quality respectively with well-defined metrics.  To evaluate the precision of attribute control, we employ the pre-trained attribute predictors to predict the attribute labels of manipulated images and then compare them with target labels. For assure fairness, we employ a different set of attribute predictors that are excluded from those used for training. Specifically, the pose estimator is the official pre-trained model from \cite{yang2018ssr}, and the age estimator is officially provided by \cite{alaluf2021only}. The glasses and smile classifier is a multi-task ResNet50 classifier \cite{he2016deep} trained by ourselves with CelebA dataset \cite{liu2015faceattributes}. With the predicted labels, we compute the Mean Absolute Error (MAE) of yaw and age, and the classification accuracy of ``glasses'' and ``smile'' attributes. As for identity preservation, we calculate the average cosine similarity score of manipulated faces and original ones: see Table \ref{table:comp}. In terms of image quality, we evaluate the distance between the distribution of manipulated images and that of real images (FFHQ dataset) with the Fréchet Inception Distance (FID)~\cite{heusel2017gans}: see Table \ref{table:comp_fid}.  As shown, our model exhibits higher control precision of all attributes except for age. When joint manipulating multiple attributes, the control precision and identity similarity scores of compared methods deteriorate significantly while our method performs consistently well. In addition, we cancel the identity loss as the ablation experiment, in order to ensure consistency with the evaluation methods, see Table.\ref{table:comp}. Although the quantitative index is slightly lower than using the identity constraint, it still has the best similarity compared with other approaches. 
	\begin{figure}[t]
		\includegraphics[width=\columnwidth]{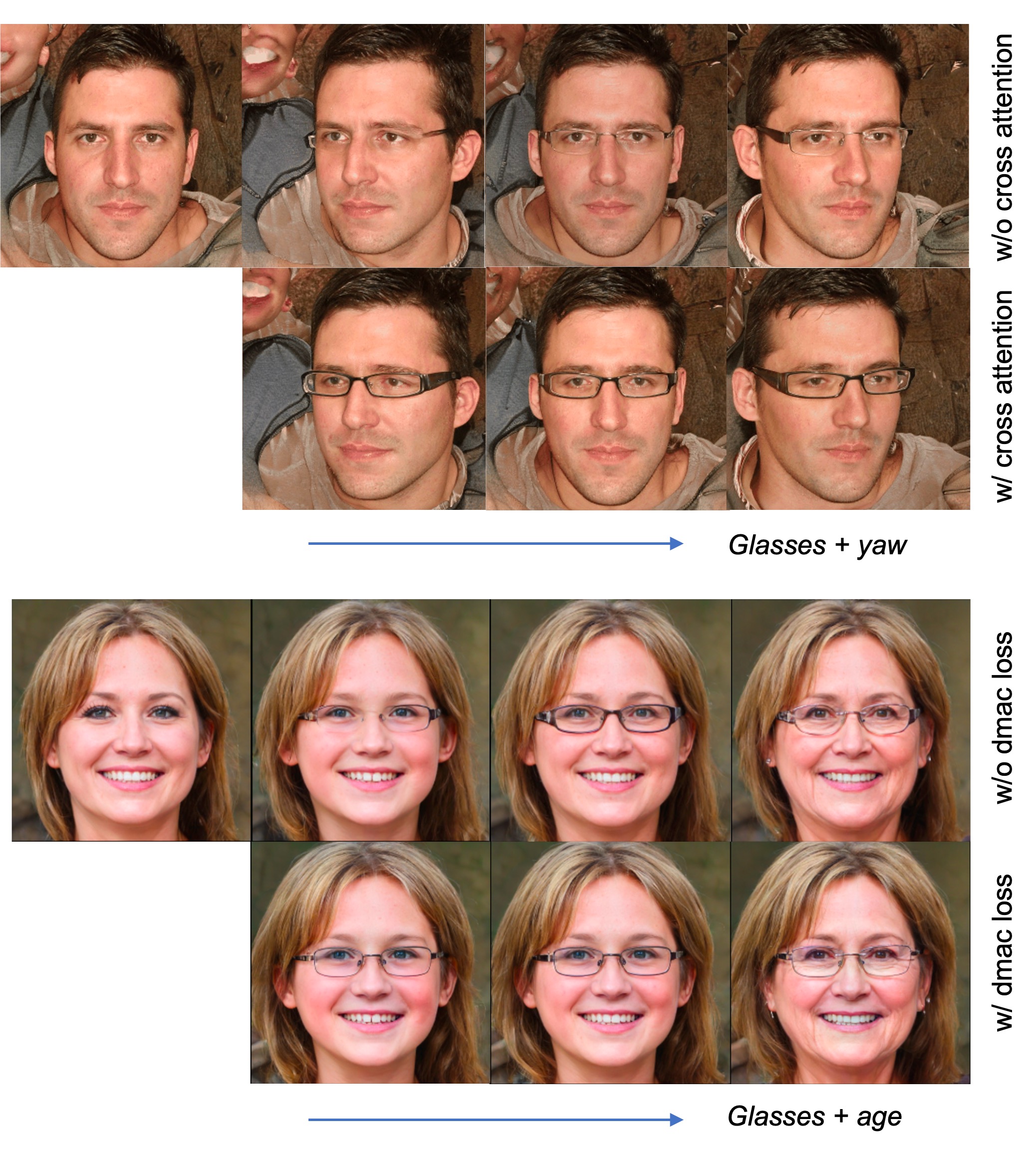}
		\centering
		\caption{Qualitatively demonstrate the stability of binary attribute under the condition of continuous editing of numerical attributes.}
		\label{fig:ab1}
	\end{figure}
	
	\setlength{\tabcolsep}{6.2pt}
	\begin{table}[h]
		\begin{center}
			\caption{The evaluation of ablation controls the accuracy of individual attribute of glasses, as well as the accuracy of the attribute of glasses under the condition of editing yaw or age.}
			\label{table:ab_1}
			\begin{tabular}{l|c|c|c}
				\hline\noalign{\smallskip}
				\multirow{2}*{Model} & \multicolumn{3}{c}{Attribute Control Accuracy($\uparrow$)}\\
				\cline{2-4} 
				\noalign{\smallskip}
				~ & glasses & glasses$|$yaw& glasses$|$age\\
				\hline
				\noalign{\smallskip}
				w/o CA \& $L_{dmac}$ & 0.943 & 0.921 & 0.919\\
				w/o CA & 0.975 & 0.947 & 0.936\\
				w/o $L_{dmac}$ & 0.951 & 0.932 & 0.924\\
				\hline
				\noalign{\smallskip}
				\bf full & \bf 0.988 & \bf 0.965 & \bf 0.96\\
				\hline
			\end{tabular}
		\end{center}
	\end{table}
	\setlength{\tabcolsep}{1.4pt}
	
	\subsection{Extended ablations studies}
	\label{sect:ab}
	
	We qualitatively and quantitatively analyze and compare ways of multi-attribute fusion (cross-attention instead of MLP) and the performance of $L_{dmac}$ loss on multi-attribute joint editing. As shown in Figure 10 and Table \ref{table:ab_1}, both tricks independently contribute to the stability of multi-attribute editing.
	
	To verify the effectiveness of the proposed dynamic architecture and two-stage training procedure, we prepared two validation datasets separately for multi-attribute-conditioned realistic (FFHQ) editing and comic face editing. We started three trainings, which include the static architecture (all branches are activated regardless of the input as in Figure \ref{fig:dnn}) trained for joint multi-attribute editing, the dynamic architecture trained with the two-stage training procedure, and that trained for multi-attribute editing only (single-stage).  We visualize how the validation losses ($L_{id}$, $L_{attr}$, $L^{A_i}_{attr}$ as defined in the method section) change against the number of training steps. As shown in Figure 8 in the supplementary materials, for both experiments, the static architecture cannot converge as well as the dynamic architecture, implying its invulnerability in adapting to various kinds of attribute configurations. As for the dynamic architecture, the single-stage training procedure is highly unstable and achieves worse identity preservation and average control accuracy. Visual comparisons are illustrated in Figures 9, Figure 10 in the supplementary materials. More ablation studies on the architecture features and training techniques are presented in the supplementary materials.
	
	\subsection{User Study}
	
	\begin{figure}[h]
		\includegraphics[width=\columnwidth]{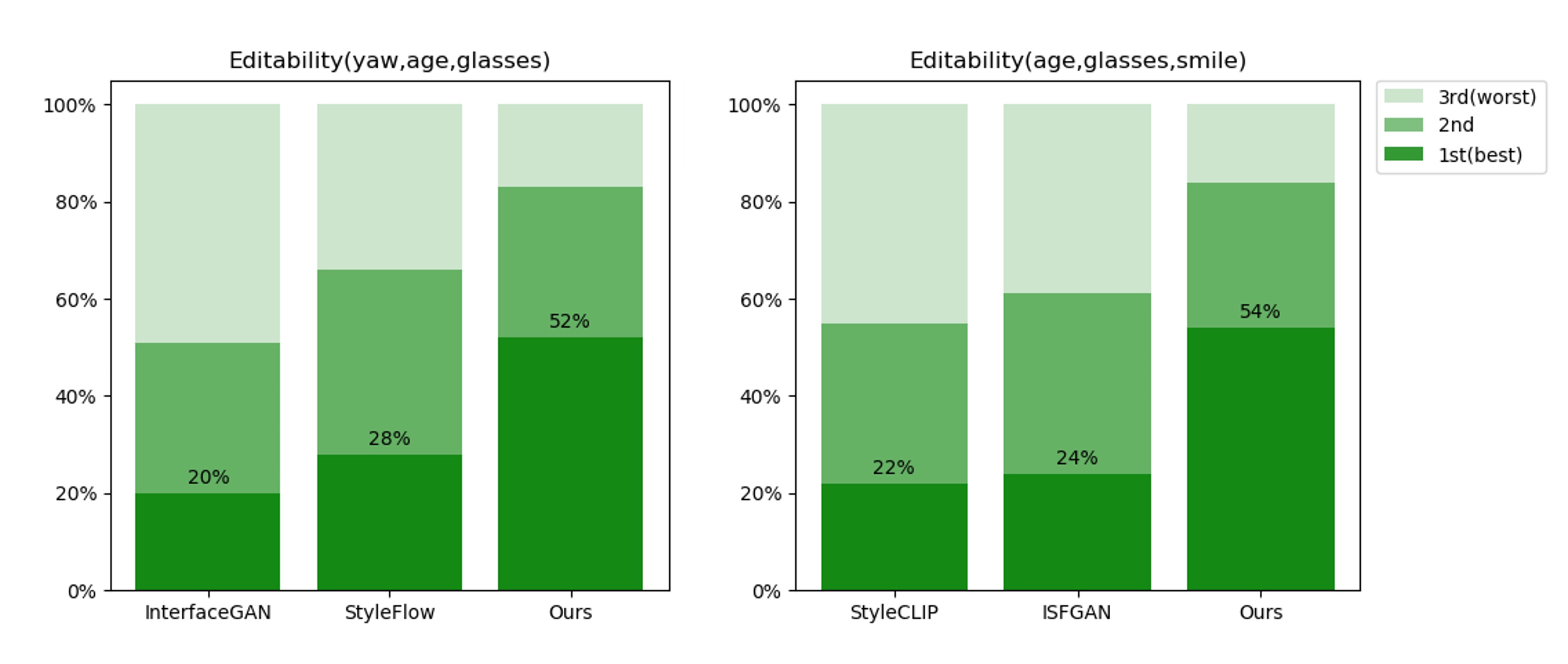}
		\centering
		\caption{User study of jointly manipulating multi-attributes, corresponding to qualitative comparisons.}
		\label{fig:user}
	\end{figure}
	To perceptually evaluate multi-attribute editing performance, we conducted a user study. We evaluate the sequential editing results of two sets of multiple attributes: the first set is yaw, age and glasses, where the presence of pose attribute allows us to compare only InterfaceGAN \cite{shen2020interfacegan} and StyleFlow \cite{abdal2021styleflow}; the second set is age, glasses, smile, we evaluate StyleCLIP  \cite{patashnik2021styleclip} and ISFGAN \cite{liu2021isf}. We collected 600 votes from 20 participants to evaluate each set of 30 images. In conclusion, our method received more than half of the first approvals.

	\section{Conclusion}
	
	In this paper, we propose a dynamic neural network that enables nonlinear and adaptive style manipulations for multi-attribute conditioned image generation. Additionally, our dynamic multi-attribute contrastive learning method effectively solves the entanglement problem of multi-attribute joint editing. Compared with other static style manipulation approaches, our model exhibits higher average precision of attribute-control and improved competency of identity preservation. When manipulating multiple attributes, the superiority of our approach becomes more significant. Future work should focus on more fine-grained division of model latent space in order to ensure the stable joint editing of more attributes.

	{\small
		\bibliographystyle{ieee_fullname}
		\bibliography{egbib}

\begin{thebibliography}{10}\itemsep=-1pt

\bibitem{blanz1999morphable}
Volker Blanz and Thomas Vetter.
\newblock A morphable model for the synthesis of 3d faces.
\newblock In {\em SIGGRAPH}, pages 187--194, 1999.

\bibitem{branwen2015danbooru2019}
Gwern Branwen.
\newblock Danbooru2019: A large-scale crowdsourced and tagged anime
  illustration dataset.
\newblock 2015.

\bibitem{alaluf2021matter}
Alaluf et al.
\newblock Only a matter of style: Age transformation using a style-based
  regression model.
\newblock {\em TOG}, 40(4), 2021.

\bibitem{alaluf2021only}
Alaluf et al.
\newblock Only a matter of style: Age transformation using a style-based
  regression model.
\newblock {\em arXiv preprint}, 2021.

\bibitem{abdal2021styleflow}
Abdal et al.
\newblock Styleflow: Attribute-conditioned exploration of stylegan-generated
  images using conditional continuous normalizing flows.
\newblock {\em TOG}, 40(3):1--21, 2021.

\bibitem{brock2018large}
Brock et al.
\newblock Large scale gan training for high fidelity natural image synthesis.
\newblock {\em arXiv preprint}, 2018.

\bibitem{choi2020stargan}
Choi et al.
\newblock Stargan v2: Diverse image synthesis for multiple domains.
\newblock In {\em CVPR}, pages 8188--8197, 2020.

\bibitem{goodfellow2014generative}
Goodfellow et al.
\newblock Generative adversarial networks.
\newblock {\em arXiv preprint}, 2014.

\bibitem{he2016deep}
He et al.
\newblock Deep residual learning for image recognition.
\newblock In {\em CVPR}, pages 770--778, 2016.

\bibitem{heusel2017gans}
Heusel et al.
\newblock Gans trained by a two time-scale update rule converge to a local nash
  equilibrium.
\newblock {\em arXiv preprint}, 2017.

\bibitem{huang2020curricularface}
Huang et al.
\newblock Curricularface: adaptive curriculum learning loss for deep face
  recognition.
\newblock In {\em CVPR}, pages 5901--5910, 2020.

\bibitem{karras2017progressive}
Karras et al.
\newblock Progressive growing of gans for improved quality, stability, and
  variation.
\newblock {\em arXiv preprint}, 2017.

\bibitem{karras2019style}
Karras et al.
\newblock A style-based generator architecture for generative adversarial
  networks.
\newblock In {\em CVPR}, pages 4401--4410, 2019.

\bibitem{karras2020analyzing}
Karras et al.
\newblock Analyzing and improving the image quality of stylegan.
\newblock In {\em CVPR}, pages 8110--8119, 2020.

\bibitem{karras2020training}
Karras et al.
\newblock Training generative adversarial networks with limited data.
\newblock {\em arXiv preprint}, 2020.

\bibitem{liu2015faceattributes}
Liu et al.
\newblock Deep learning face attributes in the wild.
\newblock In {\em ICCV}, December 2015.

\bibitem{oord2018representation}
Oord et al.
\newblock Representation learning with contrastive predictive coding.
\newblock {\em arXiv preprint}, 2018.

\bibitem{patashnik2021styleclip}
Patashnik et al.
\newblock Styleclip: Text-driven manipulation of stylegan imagery.
\newblock {\em arXiv preprint}, 2021.

\bibitem{rothe2015dex}
Rothe et al.
\newblock Dex: Deep expectation of apparent age from a single image.
\newblock In {\em ICCV workshops}, pages 10--15, 2015.

\bibitem{radford2015unsupervised}
Radford et al.
\newblock Unsupervised representation learning with deep convolutional
  generative adversarial networks.
\newblock {\em arXiv preprint}, 2015.

\bibitem{richardson2020encoding}
Richardson et al.
\newblock Encoding in style: a stylegan encoder for image-to-image translation.
\newblock {\em arXiv preprint}, 2020.

\bibitem{shen2020interfacegan}
Shen et al.
\newblock Interfacegan: Interpreting the disentangled face representation
  learned by gans.
\newblock {\em TPAMI}, 2020.

\bibitem{tewari2020stylerig}
Tewari et al.
\newblock Stylerig: Rigging stylegan for 3d control over portrait images.
\newblock In {\em CVPR}, pages 6142--6151, 2020.

\bibitem{tov2021designing}
Tov et al.
\newblock Designing an encoder for stylegan image manipulation.
\newblock {\em TOG}, 40(4):1--14, 2021.

\bibitem{wang2021hijack}
Wang et al.
\newblock Hijack-gan: Unintended-use of pretrained, black-box gans.
\newblock In {\em CVPR}, pages 7872--7881, 2021.

\bibitem{yang2018ssr}
Yang et al.
\newblock Ssr-net: A compact soft stagewise regression network for age
  estimation.
\newblock In {\em IJCAI}, volume~5, page~7, 2018.

\bibitem{zhang2018unreasonable}
Zhang et al.
\newblock The unreasonable effectiveness of deep features as a perceptual
  metric.
\newblock In {\em CVPR}, pages 586--595, 2018.

\bibitem{harkonen2020ganspace}
Erik et~al. H{\"a}rk{\"o}nen.
\newblock Ganspace: Discovering interpretable gan controls.
\newblock {\em arXiv preprint arXiv:2004.02546}, 2020.

\bibitem{Karras2020ada}
Tero Karras, Miika Aittala, Janne Hellsten, Samuli Laine, Jaakko Lehtinen, and
  Timo Aila.
\newblock Training generative adversarial networks with limited data.
\newblock In {\em Proc. NeurIPS}, 2020.

\bibitem{Karras2021}
Tero Karras, Miika Aittala, Samuli Laine, Erik H\"ark\"onen, Janne Hellsten,
  Jaakko Lehtinen, and Timo Aila.
\newblock Alias-free generative adversarial networks.
\newblock In {\em Proc. NeurIPS}, 2021.

\bibitem{kingma2014adam}
Diederik~P Kingma and Jimmy Ba.
\newblock Adam: A method for stochastic optimization.
\newblock {\em arXiv preprint}, 2014.

\bibitem{liu2021isf}
Yahui Liu, Yajing Chen, Linchao Bao, Nicu Sebe, Bruno Lepri, and Marco
  De~Nadai.
\newblock Isf-gan: An implicit style function for high-resolution
  image-to-image translation.
\newblock {\em arXiv preprint arXiv:2109.12492}, 2021.

\bibitem{mirza2014conditional}
Mehdi Mirza and Simon Osindero.
\newblock Conditional generative adversarial nets.
\newblock {\em arXiv preprint}, 2014.

\bibitem{park2019SPADE}
Taesung Park, Ming-Yu Liu, Ting-Chun Wang, and Jun-Yan Zhu.
\newblock Semantic image synthesis with spatially-adaptive normalization.
\newblock In {\em Proceedings of the IEEE Conference on Computer Vision and
  Pattern Recognition}, 2019.

\bibitem{2021Learning}
A. Radford, J.~W. Kim, C. Hallacy, A. Ramesh, G. Goh, S. Agarwal, G. Sastry, A.
  Askell, P. Mishkin, and J. Clark.
\newblock Learning transferable visual models from natural language
  supervision.
\newblock 2021.

\bibitem{ruiz2018fine}
Nataniel et~al. Ruiz.
\newblock Fine-grained head pose estimation without keypoints.
\newblock In {\em CVPR workshops}, pages 2074--2083, 2018.

\bibitem{shen2020closed}
Yujun Shen and Bolei Zhou.
\newblock Closed-form factorization of latent semantics in gans.
\newblock {\em arXiv preprint}, 2020.

\bibitem{shi2022semanticstylegan}
Yichun Shi, Xiao Yang, Yangyue Wan, and Xiaohui Shen.
\newblock Semanticstylegan: Learning compositional generative priors for
  controllable image synthesis and editing.
\newblock In {\em Proceedings of the IEEE/CVF Conference on Computer Vision and
  Pattern Recognition}, pages 11254--11264, 2022.

\bibitem{wang2021sketch}
Sheng-Yu Wang, David Bau, and Jun-Yan Zhu.
\newblock Sketch your own gan.
\newblock In {\em Proceedings of the IEEE/CVF International Conference on
  Computer Vision}, pages 14050--14060, 2021.

\bibitem{xia2021tedigan}
Weihao Xia, Yujiu Yang, Jing-Hao Xue, and Baoyuan Wu.
\newblock Tedigan: Text-guided diverse face image generation and manipulation.
\newblock In {\em Proceedings of the IEEE/CVF conference on computer vision and
  pattern recognition}, pages 2256--2265, 2021.

\bibitem{zeng2022sketchedit}
Yu Zeng, Zhe Lin, and Vishal~M Patel.
\newblock Sketchedit: Mask-free local image manipulation with partial sketches.
\newblock In {\em Proceedings of the IEEE/CVF Conference on Computer Vision and
  Pattern Recognition}, pages 5951--5961, 2022.

\end{thebibliography}
	}
	
	\clearpage

	\section{Appendix}
	
	\subsection{Extended experiment setups}
	As the training of our style manipulation network does not require any external images, the training samples are basically randomly-sampled attribute-configuration-and-$\textbf{W}+$ pairs, which are generated on the fly during training. Specifically, the $\textbf{W}+$ vectors are mapped from randomly-sampled gaussian noise vector $\textbf{z}\sim N(\textbf{0},\textbf{1})$ (truncated with 0.7) with the style mapping network of StyleGAN2 \cite{karras2020analyzing}, and the attribute configurations are produced by evenly sampling the attribute values along each attribute space (e.g., $age\sim U[-30,30]$, $smile\sim U\{0,1\}$). The detailed attribute settings in our experiments for different datasets are listed in Table \ref{table:att_set}.
	
	As shown in Figure \ref{fig:framework}, we employ a couple of pre-trained attribute prediction networks to supervise the training. Specifically, we employ the official pre-trained HopeNet model \cite{ruiz2018fine} implemented in PyTorch for pose estimation. The age estimator used for training is the pre-trained age regressor implemented in PyTorch \cite{rothe2015dex}. We employ the official pre-trained CircularFace model \cite{huang2020curricularface} for identity embedding extraction of realistic/artistic faces. The official pre-trained VGG-19 model for comic/animal identity embedding extraction. The multi-task binary attribute classifier is a ResNet34 \cite{he2016deep} trained by ourselves on the CelebA dataset~\cite{liu2015faceattributes} (for realistic/artistic faces), AFHQ (for animal faces), and comics dataset \cite{branwen2015danbooru2019} (for comic face).
	
	Our style manipulation network is implemented in PyTorch 1.6. It is trained with batch size of 8 on a single Tesla V100 GPU. It is optimized using Adam optimizer \cite{kingma2014adam} with $\beta_1$ = 0.5 and $\beta_2$ = 0.99, and the learning rate is fixed at $10^{-4}$. In all experiments, our model is trained for 50,000 steps for single-attribute manipulation (Stage I) and then 100,000 additional epochs for for multi-attribute editing (Stage II).

	\subsection{Extended ablations studies}
	\label{sect:ab}
	Other than the dynamic architecture, $L_{dmac}$ constraint and two-stage training procedure, we conducted more ablation studies on other architecture features and training techniques used in our approach.
	
	\noindent\textbf{Relative numeric attribute} In our design, the numeric attribute (e.g., age, yaw, pitch) values on which the style manipulation network is conditioned represent the relative change of the attribute with respect to the original image, rather than the absolute attribute values. For example, if the user specified  $yaw=+15°$, it means increasing the yaw angle by 15° with respect to the original face (i.e., $\Delta yaw=+15°$).  Such relative feature for numeric attributes has proved to be more effective in encouraging precise edits. Figure \ref{fig:ab2} demonstrates if the relative attribute setting (only for numeric attributes) and the use of the contrastive losses is superior to the absolute attribute setting.
	
	\noindent\textbf{Additional architectural details} Figure \ref{fig:ab4} examines how our model benefits from the identity loss $L_{id}$ and normalization loss $L_{norm}$ as introduced in the method section.  Figure \ref{fig:ab3} compares two variations of DyStyle architectures. In our architecture design, we condition the proxy codes on both the latent code $\textbf{W}_l+$ and attribute specification to allow adaptive style modulation. Whereas, an alternative way is to condition the generation of proxy codes merely on the attributes. Figure \ref{fig:ab3} verifies the superiority of this feature.

	%
	
	\begin{figure}[t]
		\centering
		\includegraphics[width=\linewidth]{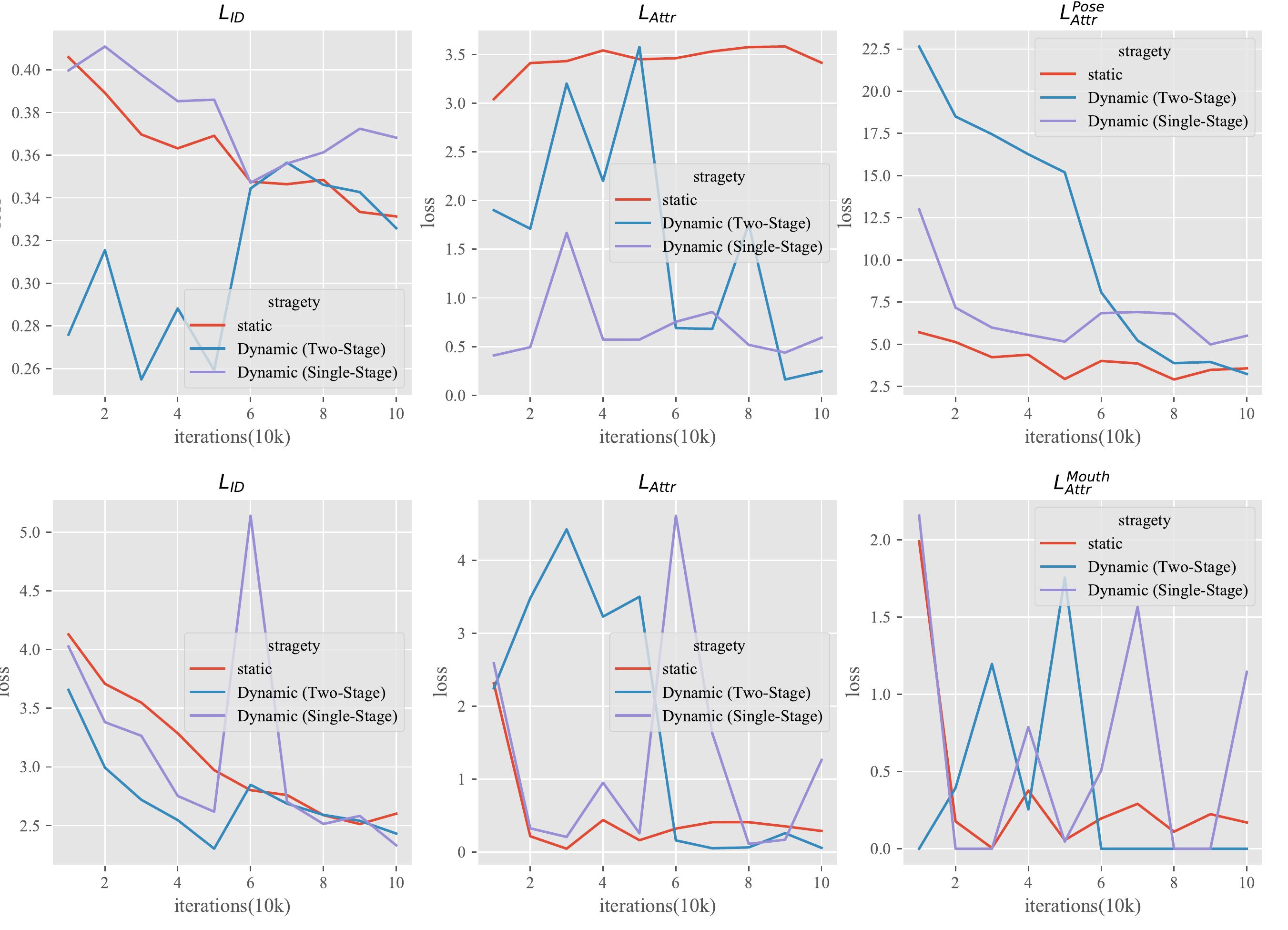}
		\caption{Comparisons of the dynamic architecture versus the static architecture, and the two-stage training versus single-stage training, experimented on FFHQ ($\textbf{top}$) and comic datasets ($\textbf{bottom}$). The validation losses demonstrate how well the model performs on the validation dataset in terms of the identity preservation and attribute control accuracy (the lower the better). As shown, as the two-stage training procedure trains for single-attribute editing in the first 50k iterations, its performance in terms of multi-attribute control accuracy at the beginning is not good. However, after 100k iterations, the two-stage training converges well and achieves the lowest $L_{id}$ and $L_{attr}$.}
		\label{fig:ab1_quant}
	\end{figure}
	
	\subsection{Visual Comparisons between DyStyle and prior methods}
	\label{sect:comp}
	\begin{table*}[]
		\centering
		\caption{The attribute settings in our experiments for different datasets.}
		\label{table:att_set}
		\begin{tabular}{l || l | l |l | l}
			& attribute name & type                                                          & value     & detailed explanations                                                                                                                                             \\
			\hline
			\hline
			
			\multicolumn{1}{c||}{\multirow{7}{*}{realistic/artistic face}} & yaw            & numeric                                                       & (-30,30)  & relative yaw change. "+20" means "increase yaw angle by 20°"                                                                                                      \\
			\multicolumn{1}{c||}{}                                         & pitch          & numeric                                                       & (-30,30)  & relative pitch change. "+20" means "increase pitch angle by 20°"                                                                                                  \\
			\multicolumn{1}{c||}{}                                         & age            & numeric                                                       & (-30,30)  & relative age change. "+20" means "become 20 years older"                                                                                                          \\
			\multicolumn{1}{c||}{}                                         & black-hair     & binary                                                        & \{0,1\}   & 1 means having black hair                                                                                                                                        \\
			\multicolumn{1}{c||}{}                                         & mustache       & binary                                                        & \{0,1\}   & 1 means having mustaache                                                                                                                                          \\
			\multicolumn{1}{c||}{}                                         & expressions    & \begin{tabular}[c]{@{}l@{}}Multi-class\\ binary\end{tabular}  & $\{0,1\}^7$  & \begin{tabular}[c]{@{}l@{}}(smile, angry, disgust, fear, sad, surprise, neutral)\\ these expressions are exclusive. 1 means having that expressions.\end{tabular} \\

			\multicolumn{1}{c||}{}                                         & glasses        & binary                                                        & \{0,1\}   & 1 means having glasses                                                                                                                                            \\
			\hline
			\multirow{4}{*}{comic face}                                  & pupil color    & \begin{tabular}[c]{@{}l@{}}Multi-class\\ binary\end{tabular}  & $\{0,1\}^8$  & \begin{tabular}[c]{@{}l@{}}(red, yellow, blue, green, brown, purple, black, white)\\ these colors are exclusive. 1 means the pupil is of that color.\end{tabular} \\
			& hair color     & \begin{tabular}[c]{@{}l@{}}Multi-class\\ binary\end{tabular}  & $\{0,1\}^8$  & \begin{tabular}[c]{@{}l@{}}(red, yellow, blue, green, brown, purple, black, white)\\ these colors are exclusive. 1 means the hair is of that color.\end{tabular}  \\
			& open mouth     & binary                                                        & \{0,1\}   & 1 means open mouth                                                                                                                                                \\
			& hair style     & \begin{tabular}[c]{@{}l@{}}Multi-class\\ binary\end{tabular}  & $\{0,1\}^2$  & \begin{tabular}[c]{@{}l@{}}(long, short).\\ the hair styles are exclusive. 1 means the hair is of that style.\end{tabular}                                        \\
			
			\hline
			
			\multirow{5}{*}{animal face}                                 & head pose      & \begin{tabular}[c]{@{}l@{}}Multi-class\\ numeric\end{tabular} & \{0,1,2\} & \{0: head turn left, 1: head facing front, 2: head turn right\}                                                                                                   \\
			& young          & binary                                                        & \{0,1\}   & 0 mens young, 1 means old                                                                                                                                         \\
			& open mouth     & binary                                                        & \{0,1\}   & 0 means open mouth. 1 means shut off the mouth.                                                                                                                   \\
			& close eye      & binary                                                        & \{0,1\}   & 1 means close eye                                                                                                                                                 \\
			& breed          & \begin{tabular}[c]{@{}l@{}}Multi-class\\ binary\end{tabular}  & $\{0,1\}^5$  & \begin{tabular}[c]{@{}l@{}}the breed set vary by dog or cat.\\ Breed types are exclusive. 1 means the cat (or dog) is of that breed.\end{tabular}            
		\end{tabular}
	\end{table*}
	We visually show some test results and demonstrate how the generated images vary by methods. As shown in Figure \ref{fig:comp}, \ref{fig:comp2}, when jointly manipulating multiple attribute, the preservation of identity and control accuracy along each attribute of prior methods are problematic. Some incremental attribute editing results are demonstrated in Figure \ref{fig:comp3}. As shown, StyleFlow \cite{abdal2021styleflow} exhibits good attribute disentanglement and identity preservation as ours. Whereas, the control precision of yaw and the smoothness of change when controlling binary attributes (glasses and smile) is inferior to ours. As InterFaceGAN \cite{shen2020interfacegan} performs linear editing of style codes, unwanted change of identity and other attributes are noticeable. StyleCLIP \cite{patashnik2021styleclip} shows good identity preservation in the process of attribute editing, but the semantic accuracy is unsatisfactory. Except for the accumulation of errors caused by static sequential editing, general CLIP \cite{2021Learning} models may not accurately describe specific attributes. ISFGAN \cite{liu2021isf} improves generative diversity but reduces stability due to the introduction of noise in the manipulation of latent codes. As all images are generated with the same StyleGAN2 generator, the degradation of image qualities is unnoticeable.  Generally speaking, he performance gap in terms of single-attribute editing between existing methods and ours are not that noticeable as joint multi-attribute editing.
	
	\subsection{Visual experimental results}
	\label{sect:visual}
	
	We present more attribute-controlled image generation results in Figure \ref{fig:example1} (realistic faces), Figure \ref{fig:example2} (artistic faces), Figure \ref{fig:example_comics} (comic faces) and Figure \ref{fig:example_animal} (animal faces). Some more high-resolution (1024$\times$1024) realistic face editing results by our approach are presented in Figure \ref{fig:example3}, in the context of single-attribute manipulation and multi-attribute manipulation. Some more high-resolution (1024$\times$1024) realistic face editing results by our approach are presented in Figure \ref{fig:example3}. With the image-to-style encoder provided in pSp \cite{richardson2020encoding}, we also conducted attribute-conditioned editing of real photos and present the results in Figure \ref{fig:photo}. 
	\begin{figure*}[t]
		\centering
		\includegraphics[width=\linewidth]{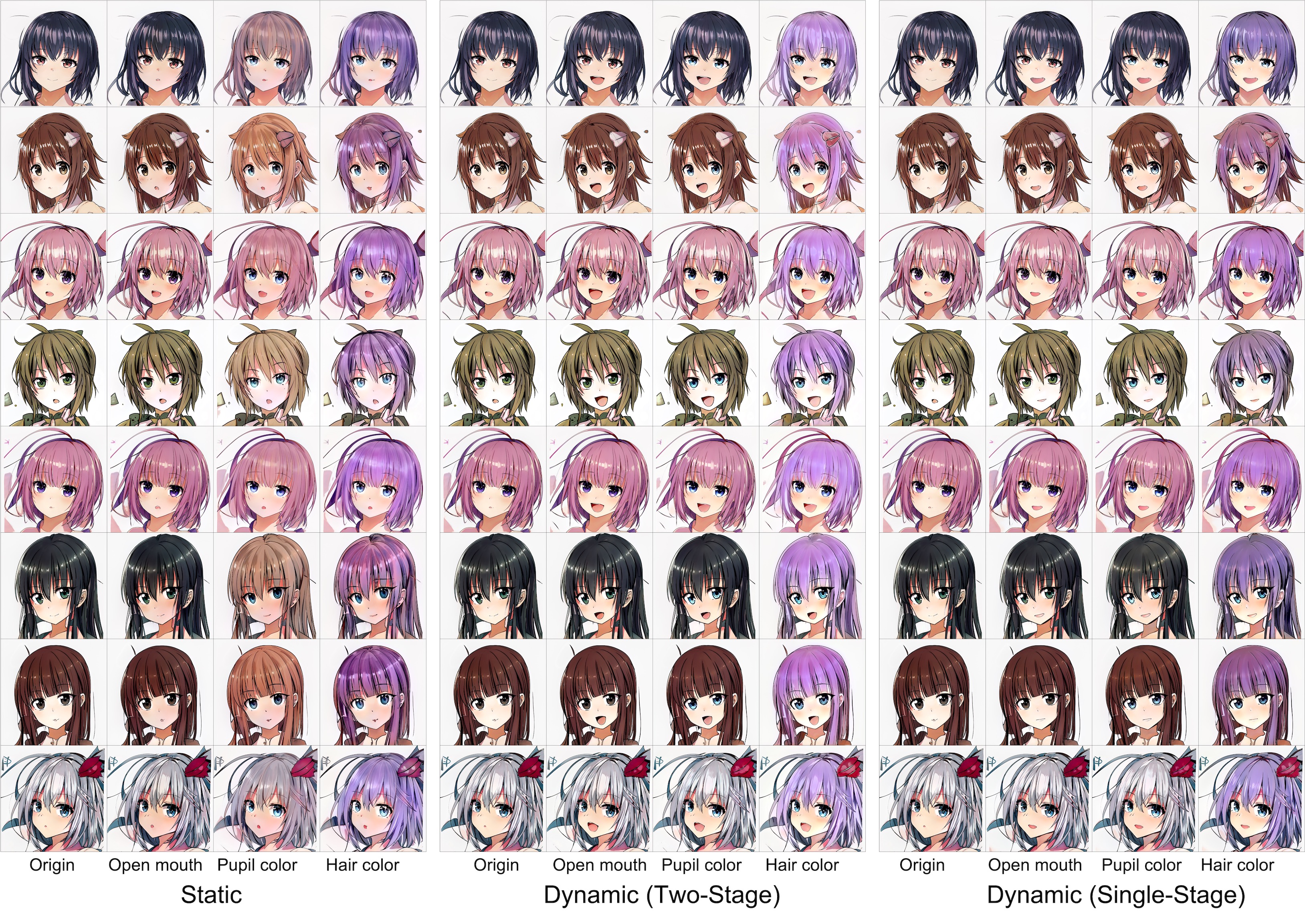}
		\caption{Qualitative comparisons of the dynamic architecture ($\textbf{middle}$) versus the static architecture ($\textbf{left}$), and the two-stage training ($\textbf{middle}$) versus single-stage training ($\textbf{right}$). As shown, static structures lead to noticeable color distortions, while single-stage dynamic structures yield weakly variable results.}
		\label{fig:ab1_qual1}
	\end{figure*}
	
	%

	\begin{figure}[t]
		\centering
		\includegraphics[width=\linewidth]{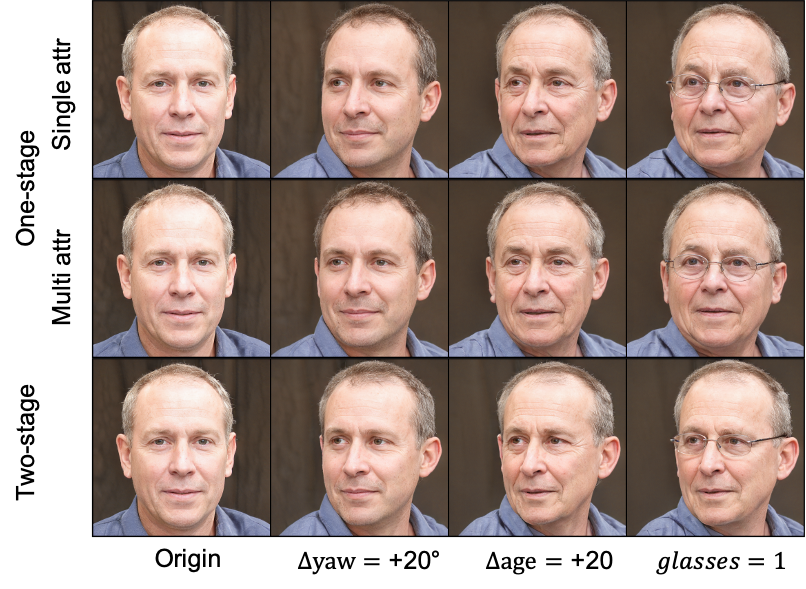}
		\caption{Ablation studies on the training strategy. We compare the results of two-stage training (single-attribute in Stage I and multi-attribute in Stage II, \textbf{(bottom)}) and one-stage training (single-attribute only \textbf{(top)}, multi-attribute only \textbf{(middle)}). As shown, the identity does not hold after multi-attribute editing \textbf{(top)} and the control of yaw is imprecise \textbf{(middle)}. The two-stage training strategy results in better editing results than one-stage training \textbf{(bottom)}.}
		\label{fig:ab1}
	\end{figure}

	\begin{figure}[t]
		\centering
		\includegraphics[width=\linewidth]{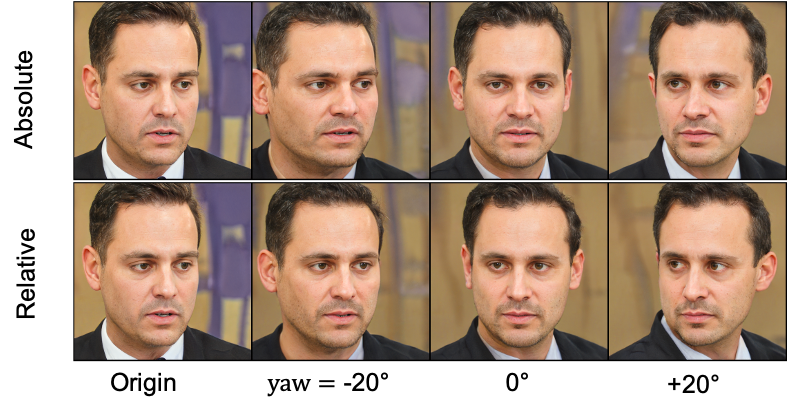}
		\vspace*{-5mm}
		\caption{Comparisons of the relative attribute setting and the absolute setting. As shown, the absolute attribute setting results in unpleasant identity variation and imprecise control of head rotation along yaw.}
		\label{fig:ab2}
	\end{figure}

	\begin{figure}
		\centering
		\includegraphics[width=\linewidth]{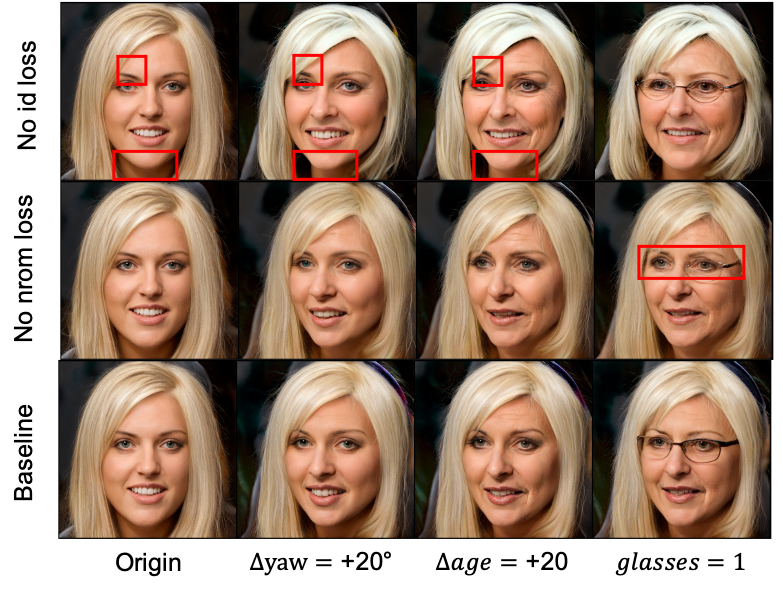}
		\caption{Ablation studies on the loss configuration. By removing the identity loss term or the normalization loss term from the full loss as in Eq \ref{eq:loss}, we retrain our model with the same hyper-parameters. Without $L_{id}$ loss, the identity variation tends to be more significant \textbf{(top)}. Without $L_{norm}$, the generated images are prone to fall into failure modes \textbf{(middle)}: see the regions highlighted with red boxes.}
		\label{fig:ab3}
	\end{figure}
	
	\begin{figure}
		\centering
		\includegraphics[width=\linewidth]{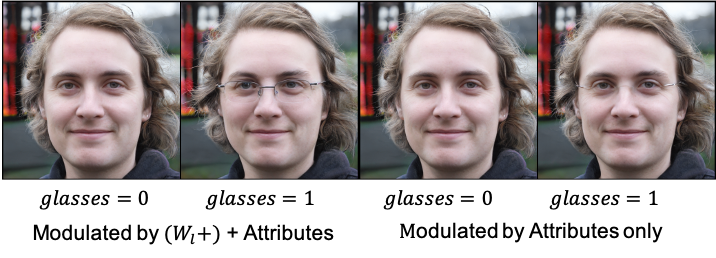}
		\vspace*{-5mm}
		\caption{Ablation studies on the architecture design of the DyStyle. We compared the architecture conditioned the generation of proxy codes on the latent code and attributes (left), and that conditioned on attributes only (right). When changing the face from ``no-glasses'' to ``with-glasses'', the left model generates faithful attribute editing results while the right model is prone to fail.}
		\label{fig:ab4}
	\end{figure}
	
	\begin{figure*}[t]
		\centering
		\includegraphics[width=\linewidth]{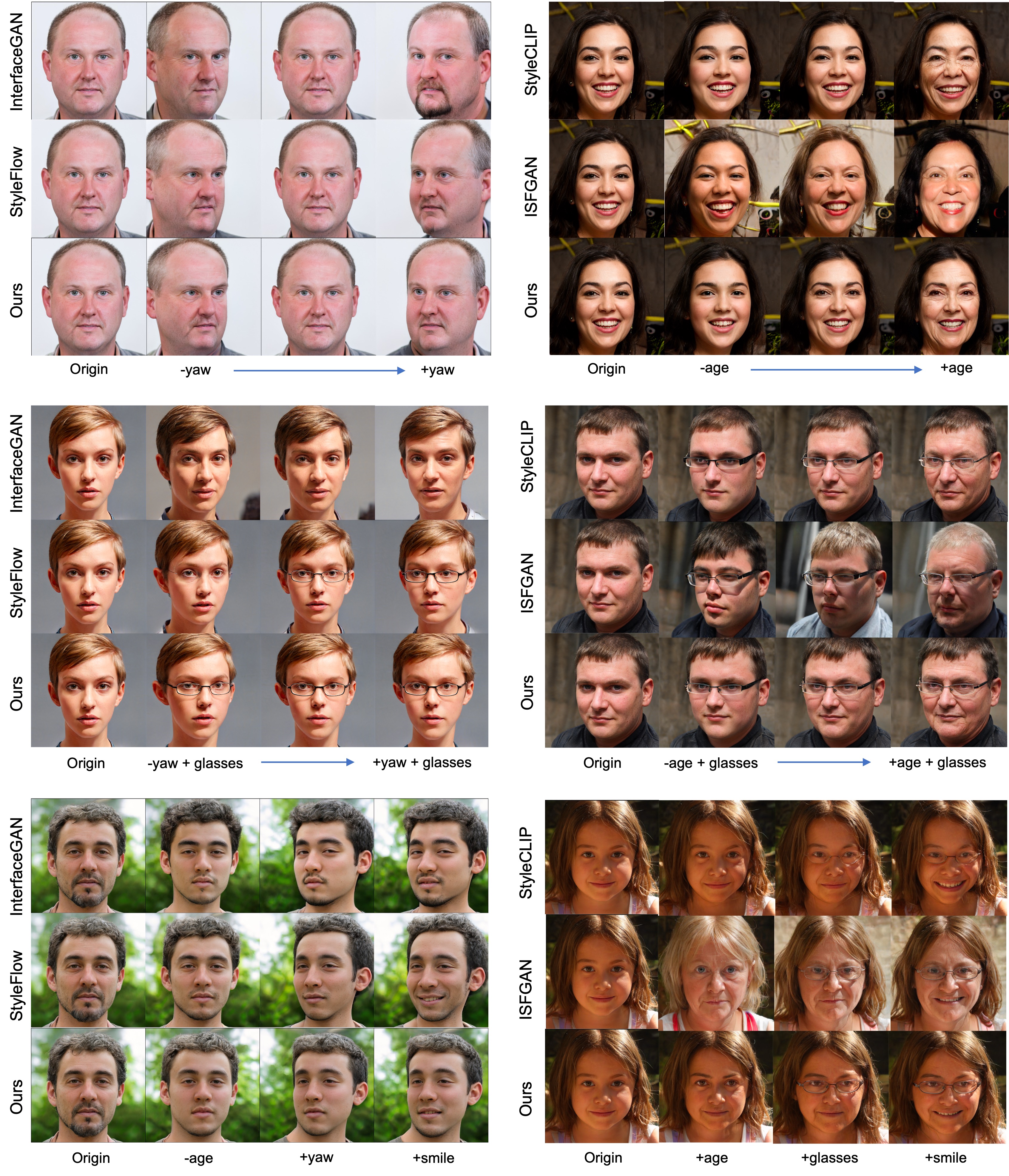}
		\caption{Comparisons of our approach and competing methods in terms of incremental attribute manipulation.}
		\label{fig:comp3}
	\end{figure*}
	
	
	
	\begin{figure*}[t]
		\centering
		\includegraphics[width=\linewidth]{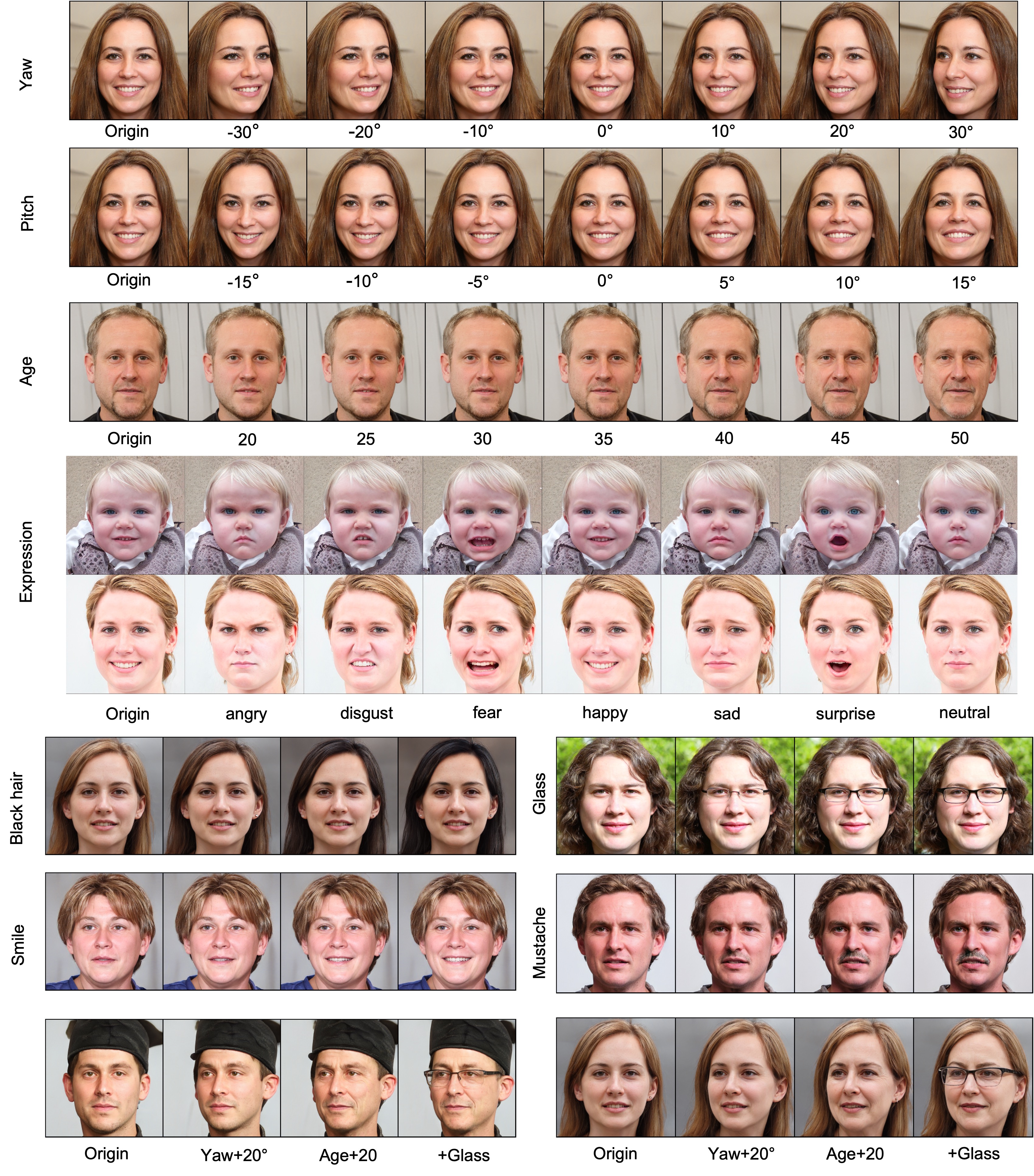}
		\caption{Results of single- and multi-attribute manipulation on realistic faces.}
		\label{fig:example1}
	\end{figure*}
	
	
	\begin{figure*}[t]
		\centering
		\includegraphics[width=\linewidth]{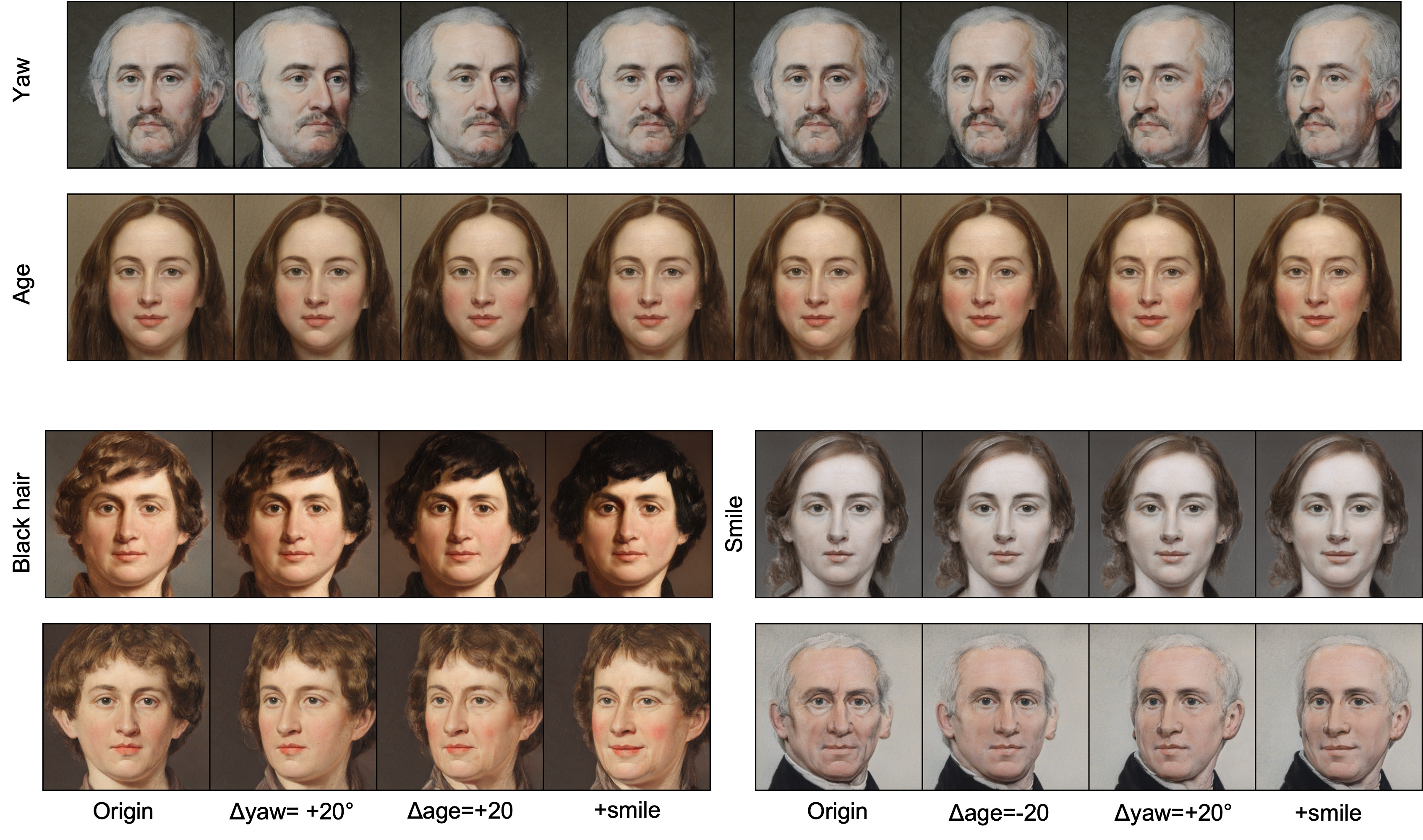}
		\caption{Results of single- and multi-attribute manipulation on artistic faces.}
		\label{fig:example2}
	\end{figure*}
	
	\begin{figure*}[t]
		\centering
		\includegraphics[width=\linewidth]{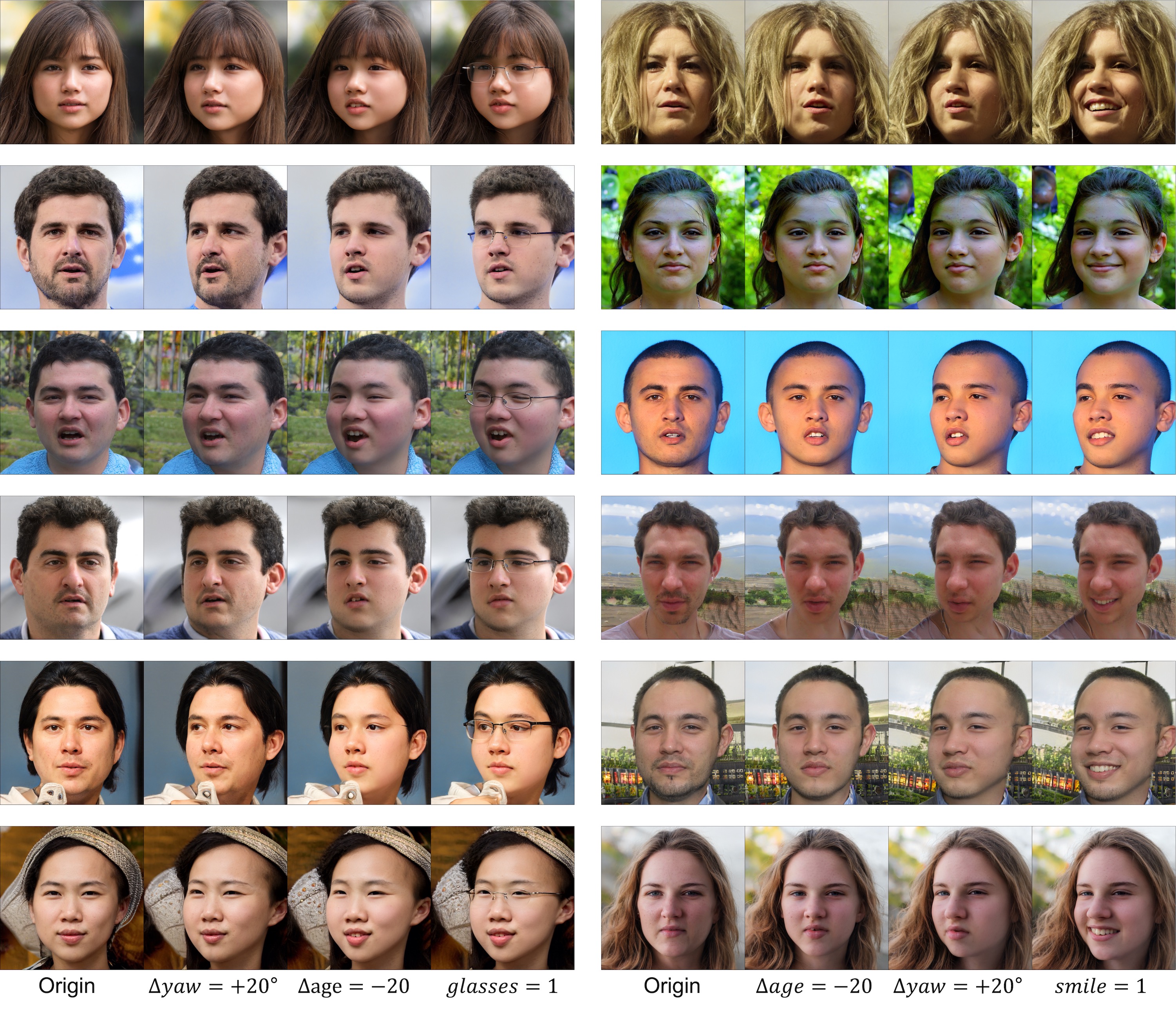}
		\caption{Results of multi-attribute manipulation on high-definition realistic faces (1024$\times$1024).}
		\label{fig:example3}
	\end{figure*}

	\begin{figure*}[t]
		\centering
		\includegraphics[width=\linewidth]{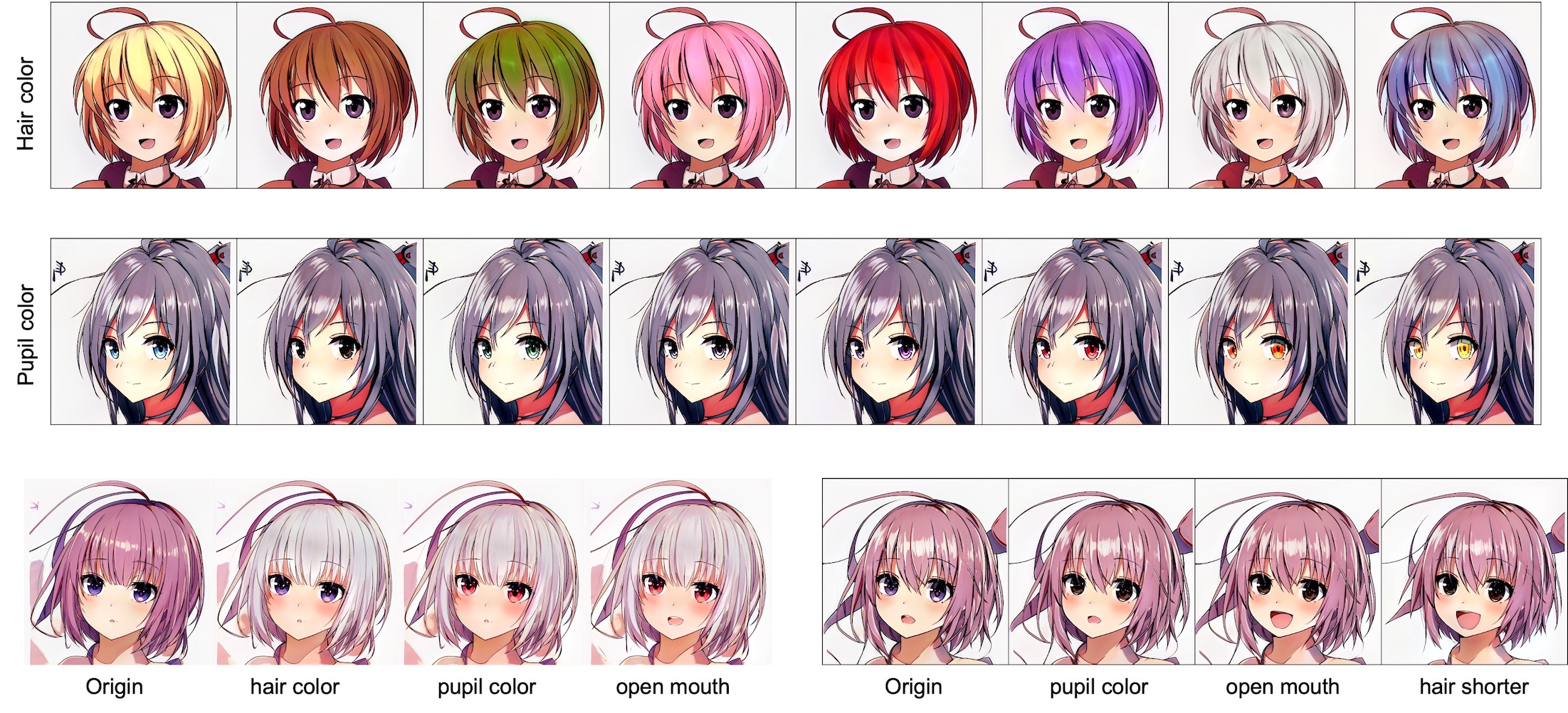}
		\caption{Results of single- and multi-attribute manipulation on comic faces.}
		\label{fig:example_comics}
	\end{figure*}
	
	\begin{figure*}[t]
		\centering
		\includegraphics[width=\linewidth]{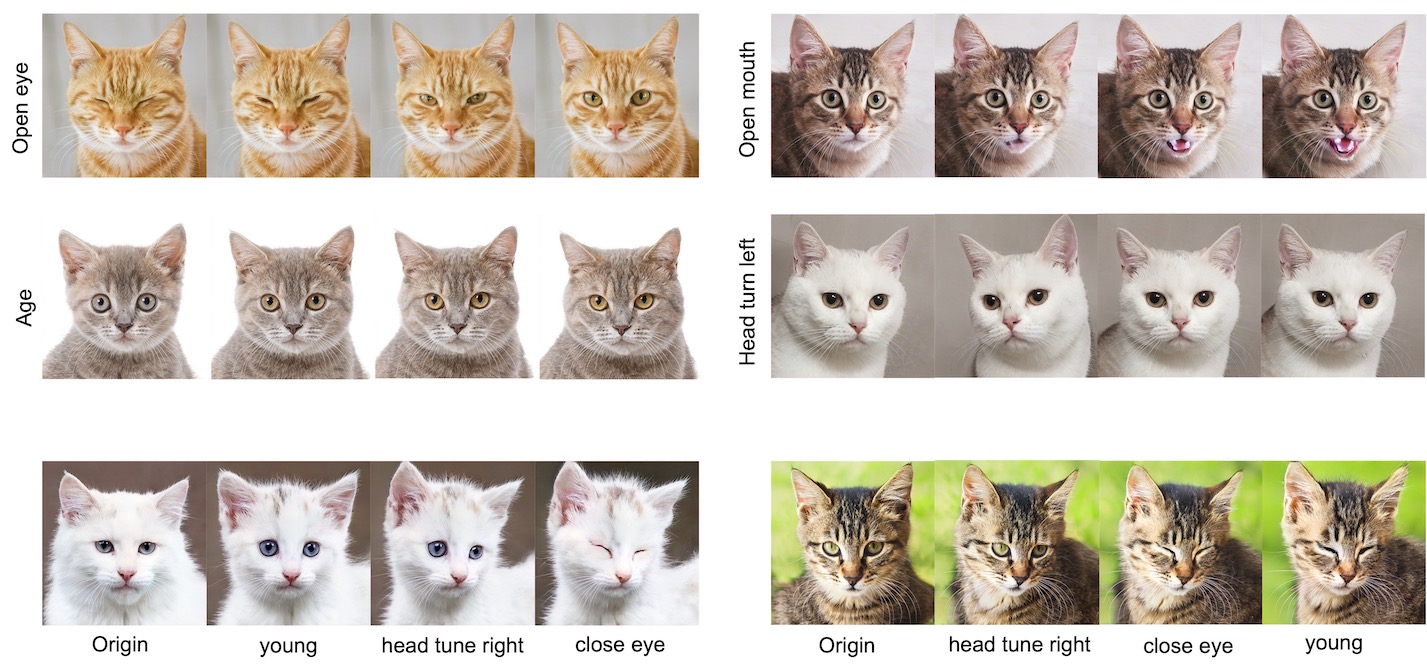}
		\caption{Results of single- and multi-attribute manipulation on animal faces.}
		\label{fig:example_animal}
	\end{figure*}
	
	\begin{figure*}[t]
		\centering
		\includegraphics[width=\linewidth]{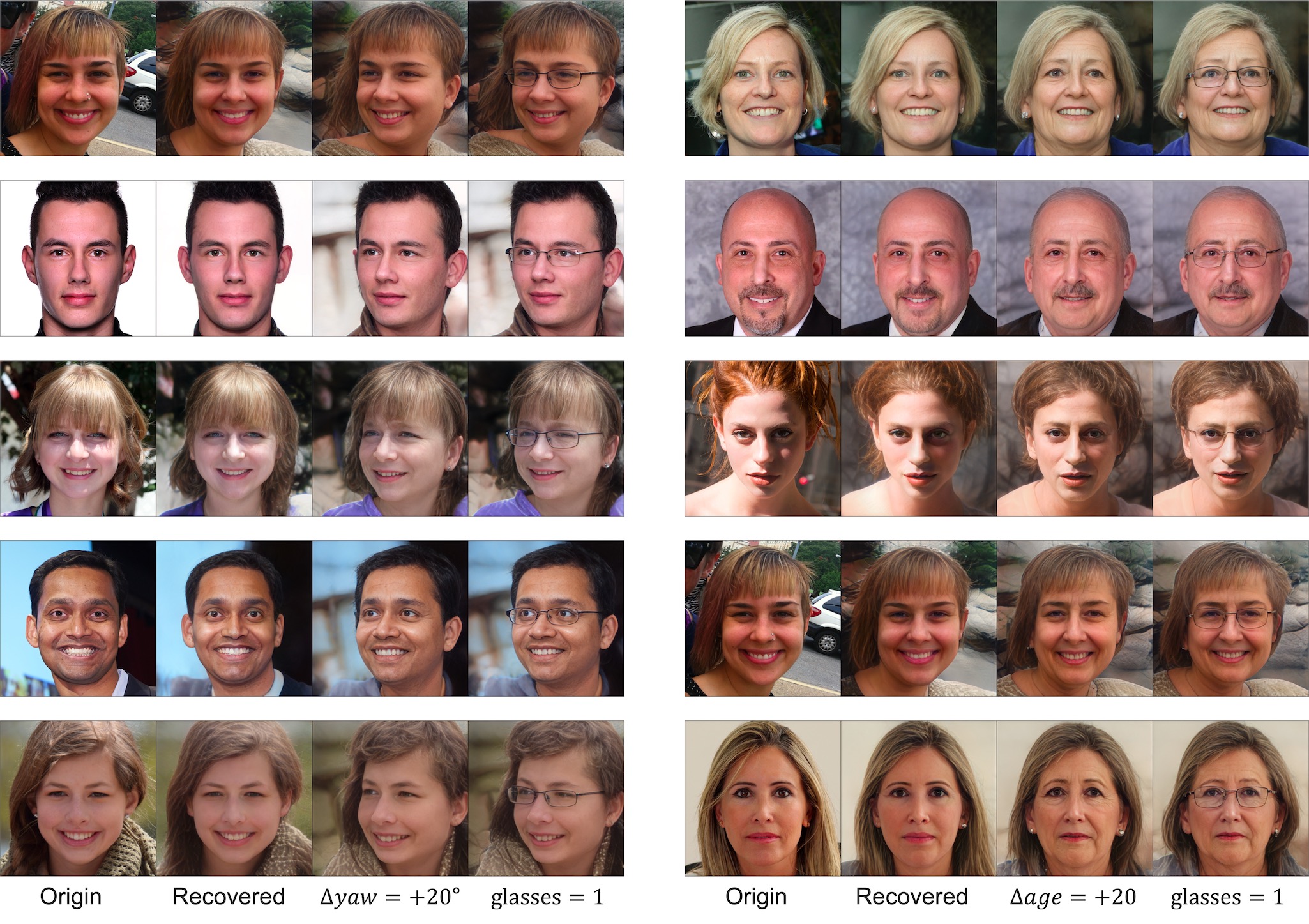}
		\caption{Multi-attribute editing of real photographs that are reconstructed with pSp encoder.}
		\label{fig:photo}
	\end{figure*}


\end{document}